\documentclass[a4paper,english]{article}
\usepackage[T1]{fontenc}
\usepackage[latin9]{inputenc}
\usepackage{verbatim}
\usepackage{float}
\usepackage{rotfloat}
\usepackage{amssymb}
\usepackage{graphicx}
\usepackage{setspace}
\usepackage[numbers]{natbib}
\PassOptionsToPackage{normalem}{ulem}
\usepackage{ulem}
\makeatletter

\providecommand{\tabularnewline}{\\}
\floatstyle{ruled}
\newfloat{algorithm}{tbp}{loa}
\providecommand{\algorithmname}{Algorithm}
\floatname{algorithm}{\protect\algorithmname}

\newcommand{\lyxaddress}[1]{
\par {\raggedright #1
\vspace{1.4em}
\noindent\par}
}

\date{}

\makeatother

\usepackage{babel}
\begin{document}

\title{Ensembles of Classifiers based on Dimensionality Reduction}

\author{Alon Schclar$^{a}$%
\thanks{Corresponding author. Tel: +972-3-6803408; Fax: +972-3-6803342%
}~ Lior Rokach$^{b}$ Amir Amit$^{c}$}

\maketitle

\lyxaddress{\begin{center}
$^{a}$School of Computer Science, Academic College of Tel Aviv-Yaffo
\\
P.O.B 8401, Tel Aviv 61083, Israel \\
alonschc@mta.ac.il 
\par\end{center}}

\lyxaddress{\begin{center}
$^{b}$Department of Information Systems Engineering, Ben-Gurion
University of the Negev \\
P.O.B 653, Beer-Sheva 84105, Israel \\
liorrk@bgu.ac.il 
\par\end{center}}

\lyxaddress{\begin{center}
$^{c}$The Efi Arazi School of Computer Science, Interdisciplinary
Center (IDC) Herzliya \\
P.O.B 167, Herzliya 46150, Israel \\
ilhostc@ilhost.com
\par\end{center}}
\begin{abstract}
We present a novel approach for the construction of ensemble classifiers
based on dimensionality reduction. Dimensionality reduction methods
represent datasets using a small number of attributes while preserving
the information conveyed by the original dataset. The ensemble members
are trained based on dimension-reduced versions of the training set.
These versions are obtained by applying dimensionality reduction to
the original training set using different values of the input parameters.
This construction meets both the diversity and accuracy criteria which
are required to construct an ensemble classifier where the former
criterion is obtained by the various input parameter values and the
latter is achieved due to the decorrelation and noise reduction properties
of dimensionality reduction. In order to classify a test sample, it
is first embedded into the dimension reduced space of each individual
classifier by using an out-of-sample extension algorithm. Each classifier
is then applied to the embedded sample and the classification is obtained
via a voting scheme. We present three variations of the proposed approach
based on the Random Projections, the Diffusion Maps and the Random
Subspaces dimensionality reduction algorithms. We also present a multi-strategy
ensemble which combines AdaBoost and Diffusion Maps. A comparison
is made with the Bagging, AdaBoost, Rotation Forest ensemble classifiers
and also with the base classifier which does not incorporate dimensionality
reduction. Our experiments used seventeen benchmark datasets from
the UCI repository. The results obtained by the proposed algorithms
were superior in many cases to other algorithms.
\end{abstract}

\section{Introduction}

Classifiers are predictive models which label data based on a training
dataset $T$ whose labels are known \emph{a-priory}. A classifier
is constructed by applying an induction algorithm, or inducer, to
$T$ - a process that is commonly known as \emph{training}. Classifiers
differ by the induction algorithms and training sets that are used
for their construction. Common induction algorithms include nearest
neighbors (NN), decision trees (CART \citep{CRT93}, C4.5 \citep{C45Quinlan}),
Support Vector Machines (SVM) \citep{SVM99} and Artificial Neural
Networks - to name a few. Since every inducer has its advantages and
weaknesses, methodologies have been developed to enhance their performance.
Ensemble classifiers are one of the most common ways to achieve that.

The need for dimensionality reduction techniques emerged in order
to alleviate the so called \emph{curse of dimensionality} \citep{JL98}.
In many cases, a high-dimensional dataset lies approximately on a
low-dimensional manifold in the ambient space. Dimensionality reduction
methods \emph{embed} datasets into a low-dimensional space while preserving
as much of the information conveyed by the dataset. The low-dimensional
representation is referred to as the \emph{embedding} of the dataset.
Since the information is inherent in the geometrical structure of
the dataset (e.g. clusters), a good embedding distorts the structure
as little as possible while representing the dataset using a number
of features that is substantially smaller than the dimension of the
original ambient space. Furthermore, an effective dimensionality reduction
algorithm also removes noisy features and inter-feature correlations.
Due to its properties, dimensionality reduction is a common step in
many machine learning applications in fields such as signal processing
\citep{SchclarDetection2010,Schclar-Neta-detection12,mine-vehicle-wave07}
and image processing \citep{Luo2012}.

\subsection{Ensembles of Classifiers\label{sub:Ensemble-Classifiers}}

Ensembles of classifiers \citep{EnsembleDiversity04} mimic the human
nature to seek advice from several people before making a decision
where the underlying assumption is that combining the opinions will
produce a decision that is better than each individual opinion. Several
classifiers (ensemble\emph{ members}) are constructed and their outputs
are combined - usually by voting or an averaged weighting scheme -
to yield the final classification \citep{Polikar,Opitz}. In order
for this approach to be effective, two criteria must be met: \emph{accuracy
}and \emph{diversity} \citep{EnsembleDiversity04}. Accuracy requires
each individual classifier to be as accurate as possible i.e. individually
minimize the generalization error. Diversity requires to minimize
the correlation among the generalization errors of the classifiers.
These criteria are contradictory since optimal accuracy achieves a
minimum and unique error which contradicts the requirement of diversity.
Complete diversity, on the other hand, corresponds to random classification
which usually achieves the worst accuracy. Consequently, individual
classifiers that produce results which are moderately better than
random classification are suitable as ensemble members. In \citep{KappaError97},
\textquotedblleft{}kappa-error\textquotedblright{} diagrams are introduced
to show the effect of diversity at the expense of reduced individual
accuracy.

In this paper we focus on ensemble classifiers that use a single induction
algorithm, for example the nearest neighbor inducer. This ensemble
construction approach achieves its diversity by manipulating the training
set. A well known way to achieve diversity is by bootstrap aggregation
(\emph{Bagging}) \citep{Bagging96}. Several training sets are constructed
by applying bootstrap sampling (each sample may be drawn more than
once) to the original training set. Each training set is used to construct
a different classifier where the repetitions fortify different training
instances. This method is simple yet effective and has been successfully
applied to a variety of problems such as spam detection \citep{BaggEnsembleApp06},
analysis of gene expressions \citep{BaggEnsembleApp03} and image
retrieval \citep{Asymmetric_Bagging_Image_Retrieval}.

The award winning Adaptive Boosting (\emph{AdaBoost}) \citep{AdaBoost96}
algorithm and its subsequent versions e.g. \citep{AdaBoost_R2} and
\citep{AdaBoost_RT} provide a different approach for the construction
of ensemble classifiers based on a single induction algorithm. This
approach iteratively assigns weights to each training sample where
the weights of the samples that are misclassified are increased according
to a global error coefficient. The final classification combines the
logarithm of the weights to yield the ensemble's classification.

Rotation Forest \citep{RotationForest2006} is one of the current
state-of-the-art ensemble classifiers. This method constructs different
versions of the training set by employing the following steps: First,
the feature set is divided into disjoint sets on which the original
training set is projected. Next, a random sample of classes is eliminated
and a bootstrap sample is selected from every projection result. Principal
Component Analysis \citep{PCA} (see Section \ref{sub:Dimensionality-reduction})
is then used to rotate each obtained subsample. Finally, the principal
components are rearranged to form the dataset that is used to train
a single ensemble member. The first two steps provide the required
diversity of the constructed ensemble.

Multi-strategy ensemble classifiers \citep{Rokach2009} aim at combining
the advantages of several ensemble algorithms while alleviating their
disadvantages. This is achieved by applying an ensemble algorithm
to the results produced by another ensemble algorithm. Examples of
this approach include multi-training SVM (MTSVM) \citep{MultitrainingLiATL06},
MultiBoosting \citep{multiboosting:a:Webb00} and its extension using
stochastic attribute selection \citep{multi-strategyensemble:Webb04}.

Successful applications of the ensemble methodology can be found in
many fields, for example, recommender systems \citep{SchclarRecSys09},
finance \citep{Leigh02-1}, manufacturing \citep{Rokach08} and medicine
\citep{Mangiameli04}.

\subsection{Dimensionality reduction\label{sub:Dimensionality-reduction} }

The theoretical foundations for dimensionality reduction were established
by Johnson and Lindenstrauss \citep{JL84} who proved its feasibility.
Specifically, they showed that $N$ points in an $N$ dimensional
space can almost always be projected onto a space of dimension $C\log N$
with control over the ratio of distances and the error (distortion).
Bourgain \citep{B85} showed that any metric space with $N$ points
can be embedded by a bi-Lipschitz map into an Euclidean space of $\log N$
dimension with a bi-Lipschitz constant of $\log N$. Various randomized
versions of these theorems were successfully applied to protein mapping
\citep{LLTY00}, reconstruction of frequency sparse signals \citep{CRT06,D06},
textual and visual information retrieval \citep{BM2001} and clustering
\citep{fern2003random}.

The dimensionality reduction problem can be formally described as
follows. Let 
\begin{equation}
\Gamma=\left\{ x_{i}\right\} _{i=1}^{N}\label{eq:training_set}
\end{equation}
be the original high-dimensional dataset given as a set of column
vectors where $x_{i}\in\mathbb{R}^{n}$, $n$ is the dimension of
the ambient space and $N$ is the size of the dataset. All dimensionality
reduction methods embed the vectors into a lower dimensional space
$\mathbb{R}^{q}$ where $q\ll n$. Their output is a set of column
vectors in the lower dimensional space 
\begin{equation}
\widetilde{\Gamma}=\left\{ \widetilde{x}_{i}\right\} _{i=1}^{N},\,\widetilde{x}_{i}\in\mathbb{R}^{q}\label{eq:gamma_gal}
\end{equation}
where $q$ is chosen such that it approximates the intrinsic dimensionality
of $\Gamma$ \citep{Hein05,Wakin07}. We refer to the vectors in the
set $\widetilde{\Gamma}$ as the \emph{embedding} \emph{vectors}.

Dimensionality reduction techniques employ two approaches: feature
selection and feature extraction. Feature selection methods reduce
the dimensionality by choosing $q$ features from the feature vectors
according to given criteria. The same features are chosen from all
vectors. Current state-of-the-art feature selection methods include,
for example, Manhattan non-negative matrix factorization \citep{ManhattanNonnegativeMatrixFactorization},
manifold elastic net \citep{Manifold_elastic_net} and geometric mean
for subspace selection \citep{GeometricMean_for_SubspaceSelection}.
Feature extraction methods, on the other hand, derive features which
are functions of the original features.

Dimensionality techniques can also be divided into \emph{global} and
\emph{local} methods. The former derive embeddings in which \emph{all}
points satisfy a given criterion. Examples for global methods include:
\begin{itemize}
\item Principal Component Analysis (PCA) \citep{PCA} which finds a low-dimensional
embedding of the data points that best preserves their variance as
measured in the ambient (high-dimensional) space; 
\item Kernel PCA (KPCA) \citep{KPCA98} which is a generalization of PCA
that is able to preserve non-linear structures. This ability relies
on the \emph{kernel trick }i.e. any algorithm whose description involves
only dot products and does not require explicit usage of the variables
can be extended to a non-linear version by using Mercer kernels \citep{KPCA_book02}.
When this principle is applied to dimensionality reduction it means
that non-linear structures correspond to linear structures in some
high-dimensional space. These structures can be detected by linear
methods using kernels. 
\item Multidimensional scaling (MDS) \citep{MDS64,MDS_94} algorithms which
find an embedding that best preserves the inter-point distances among
the vectors according to a given metric. This is achieved by minimizing
a loss/cost \emph{stress function} that measures the error between
the pairwise distances of the embedding and their corresponding distances
in the original dataset.
\item ISOMAP \citep{ISO00} which applies MDS using the \emph{geodesic distance}
metric. The geodesic distance between a pair of points is defined
as the length of the shortest path connecting these points that passes
only through points in the dataset. 
\item Random projections \citep{CRT06,D06} in which every high-dimensional
vector is projected onto a random matrix in order to obtain the embedding
vector. This method is described in details in Section \ref{sec:Random-Projections}. 
\end{itemize}
Contrary to global methods, local methods construct embeddings in
which only \emph{local} neighborhoods are required to meet a given
criterion. The global description of the dataset is derived by the
aggregation of the local\emph{ }neighborhoods. Common local methods
include Local Linear Embedding (LLE) \citep{LLE00}, Laplacian Eigenmaps
\citep{Laplacian03}, Hessian Eigenmaps \citep{Hessian02} and Diffusion
Maps \citep{CL_DM06,SchclarDiffusionFrame} which is used in this
paper and is described in Section \ref{sec:Diffusion-Maps}. The patch
alignment framework \citep{PatchAlignment_forDimensionalityReduction}
provides a unified framework to local dimensionality reduction techniques
that employ two steps: (a) an optimization step where the local criterion
is applied; and an alignment step in which the embedding is found.
Examples that fit this framework include Local Linear Embedding (LLE)
\citep{LLE00}, Laplacian Eigenmaps \citep{Laplacian03}, Hessian
Eigenmaps \citep{Hessian02}, Local tangent space alignment \citep{LocalTanSpAnal02}
and Discriminative Locality Alignment (DLA) \citep{PatchAlignment_forDimensionalityReduction}.

A key aspect of dimensionality reduction is how to efficiently embed
a \emph{new point }into a \emph{given} dimension-reduced space. This
is commonly referred to as \emph{out-of-sample extension} where the
sample stands for the original dataset whose dimensionality was reduced
and does not include the new point. An accurate embedding of a new
point requires the recalculation of the entire embedding. This is
impractical in many cases, for example, when the time and space complexity
that are required for the dimensionality reduction is quadratic (or
higher) in the size of the dataset. An efficient out-of-sample extension
algorithm embeds the new point without recalculating the entire embedding
- usually at the expense of the embedding accuracy. 

The Nyström extension \citep{N28} algorithm, which is used in this
paper, embeds a new point in linear time using the quadrature rule
when the dimensionality reduction involves eigen-decomposition of
a kernel matrix. Algorithms such as Laplacian Eigenmaps, ISOMAP, LLE,
and Diffusion Maps are examples that fall into this category and,
thus, the embeddings that they produce can be extended using the Nyström
extension \citep{kernel_view_of_DR,KPCA_view04_1}. A formal description
of the Nyström extension is given in the Sec. \ref{sec:OUT-OF-SAMPLE-EXTENSION}. 

The main contribution of this paper is a novel framework for the construction
of ensemble classifiers based on dimensionality reduction and out-of-sample
extension. This approach achieves both the diversity and accuracy
which are required for the construction of an effective ensemble classifier
and it is general in the sense that it can be used with any inducer
and any dimensionality reduction algorithm as long as it can be coupled
with an out-of-sample extension method that suits it.

The rest of this paper is organized as follows. In Section \ref{sec:The-proposed-approach}
we describe the proposed approach. In Sections \ref{sec:Diffusion-Maps},
\ref{sec:Random-Projections} and \ref{sec:Random-Subspaces} we introduce
ensemble classifiers that are based on the Diffusion Maps, random
projections and random subspaces dimensionality reduction algorithms,
respectively. Experimental results are given in Section \ref{sec:Experimental-results}.
We conclude and describe future work in Section \ref{sec:Conclusion-and-future}.

\section{Dimensionality reduction ensemble classifiers \label{sec:The-proposed-approach}}

The proposed approach achieves the diversity requirement of ensemble
classifiers by applying a given dimensionality reduction algorithm
to a given training set using different values for its input parameters.
An input parameter that is common to all dimensionality reduction
techniques is the dimension of the embedding space. In order to obtain
sufficient diversity, the dimensionality reduction algorithm that
is used should incorporate additional input parameters or, alternatively,
incorporate a randomization step. For example, the Diffusion Maps
\citep{CL_DM06} dimensionality algorithm uses an input parameter
that defines the size of the local neighborhood of a point. Variations
of this notion appear in other local dimensionality reduction methods
such as LLE \citep{LLE00} and Laplacian Eigenmaps \citep{Laplacian03}.
The Random Projections \citep{D06} (Section \ref{sec:Random-Projections})
and Random Subspaces \citep{RandomSubspaceDecisionForest,RandSubSpaceEnsemble04}
(Section \ref{sec:Random-Subspaces}) methods, on the other hand,
do not include input parameters other than the dimensionality of the
embedding space. However, they incorporate a randomization step which
diversifies the data (this approach already demonstrated good results
using Random Projections in \citep{SchclarICEIS09} and we extend
them in this paper). In this sense, PCA is not suitable for the proposed
framework since it does not include a randomization step and the only
input parameter it has is the dimension of the embedding space (this
parameter can also be set according to the total amount of variance
of the original dataset that the embedding is required to maintain).
Thus, PCA offers no way to diversify the data. On the other hand,
dimensionality reduction algorithms that are suitable for the proposed
method include ISOMAP \citep{ISO00}, LLE \uline{\mbox{\citep{LLE00}}},
Hessian LLE \citep{Hessian02}, Local tangent space alignment \citep{LocalTanSpAnal02}
and Discriminative Locality Alignment (DLA) \citep{PatchAlignment_forDimensionalityReduction}.
These methods are suitable since they require as input the number
of nearest neighbors to determine the size of the local neighborhood
of each data point. Laplacian Eigenmaps \citep{Laplacian03} and KPCA
\citep{KPCA98} are also suitable for the proposed framework as they
include a continuous input variable to determine the radius of the
local neighborhood of each point.

\begin{figure}
\centering{}\includegraphics[bb=0bp 0bp 525bp 225bp,clip,width=0.9\columnwidth]{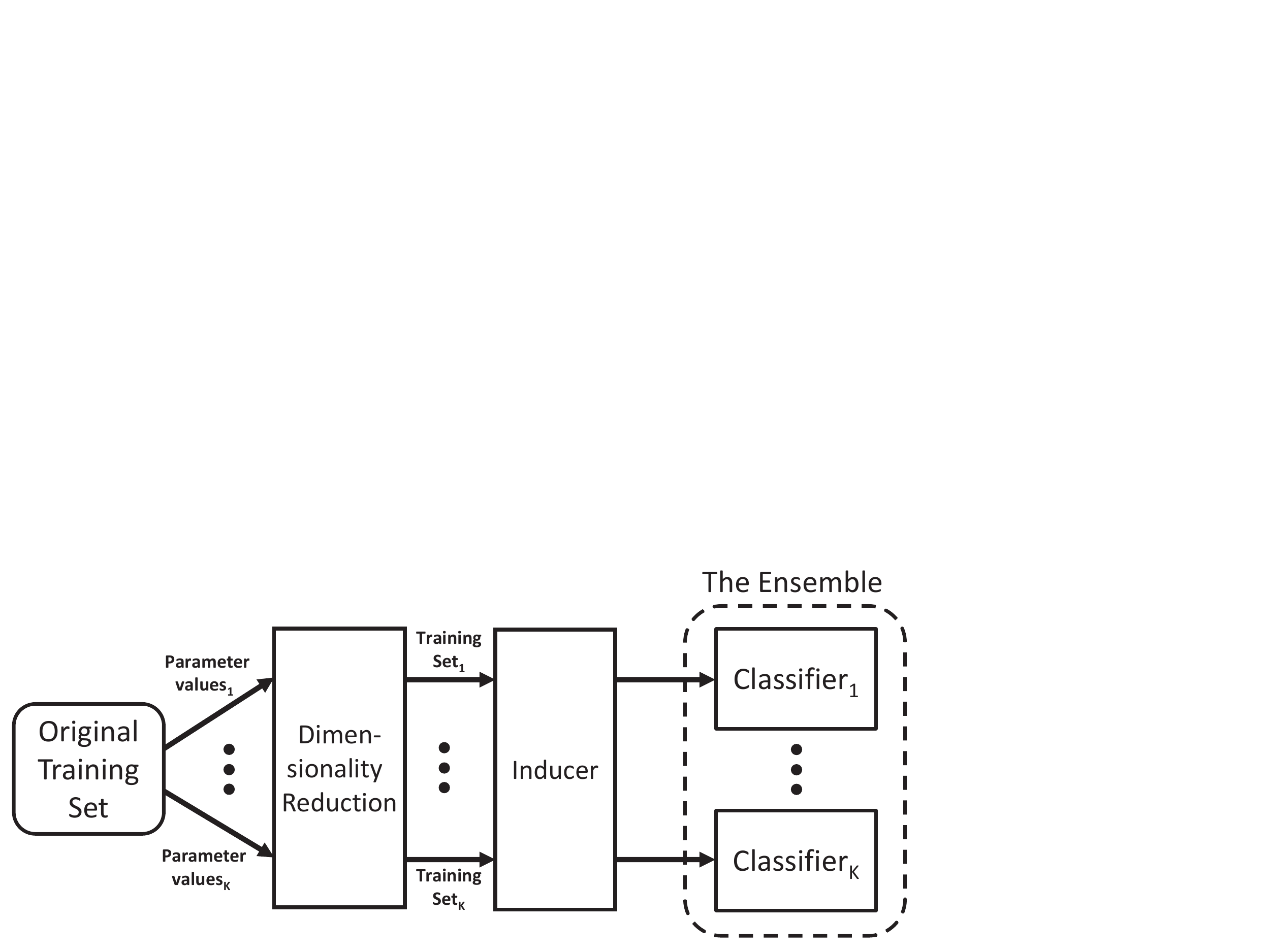}\caption{Ensemble training.\label{fig:Ensemble-training}}
\end{figure}

After the training sets are produced by the dimensionality reduction
algorithms, each set is used to train a classifier to produce one
of the ensemble members. The training process is illustrated in Fig.
\ref{fig:Ensemble-training}.

Employing dimensionality reduction to a training set has the following
advantages:
\begin{itemize}
\item It reduces noise and decorrelates the data. 
\item It reduces the computational complexity of the classifier construction
and consequently the complexity of the classification.
\item It can alleviate over-fitting by constructing combinations of the
variables \citep{Plastria}.
\end{itemize}
These points meet the accuracy and diversity criteria which are required
to construct an effective ensemble classifier and thus render dimensionality
reduction a technique which is tailored for the construction of ensemble
classifiers. Specifically, removing noise from the data contributes
to the accuracy of the classifier while diversity is obtained by the
various dimension-reduced versions of the data.

\begin{figure}
\centering{}\includegraphics[bb=0bp 5bp 555bp 265bp,clip,width=0.9\columnwidth]{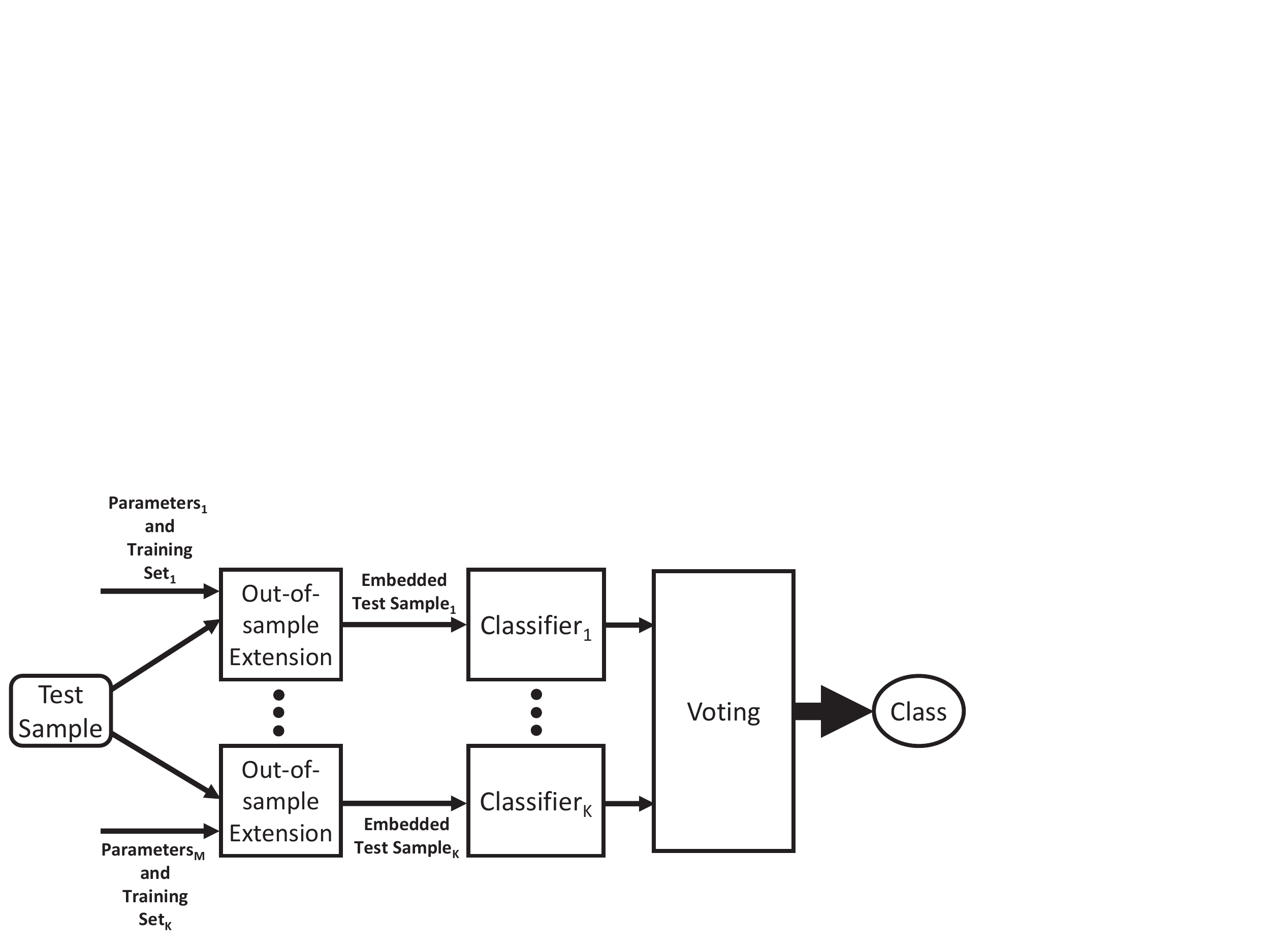}\caption{Classification process of a test sample.\label{fig:Ensemble-testing}}
\end{figure}

In order to classify test samples, they are first embedded into the
low-dimensional space of each of the training sets using out-of-sample
extension. Next, each ensemble member is applied to its corresponding
embedded test sample and the produced results are processed by a voting
scheme to derive the result of the ensemble classifier. Specifically,
each classification is given as a vector containing the probabilities
of each possible label. These vectors are aggregated and the ensemble
classification is chosen as the label with the largest probability.
Figure \ref{fig:Ensemble-testing} depicts the classification process
of a test sample.

\section{Diffusion Maps\label{sec:Diffusion-Maps}}

The Diffusion Maps (DM) \citep{CL_DM06} algorithm embeds data into
a low-dimensional space where the geometry of the dataset is defined
in terms of the connectivity between every pair of points in the ambient
space. Namely, the similarity between two points $x$ and $y$ is
determined according to the number of paths connecting $x$ and $y$
via points in the dataset. This measure is robust to noise since it
takes into account all the paths connecting $x$ and $y$. The Euclidean
distance between $x$ and $y$ in the dimension-reduced space approximates
their connectivity in the ambient space.

Formally, let $\Gamma$ be a set of points in $\mathbb{R}^{n}$ as
defined in Eq. \ref{eq:training_set}. A weighted undirected graph
$G\left(V,E\right),\,\left|V\right|=N,\,\left|E\right|\ll N^{2}$
is constructed, where each vertex $v\in V$ corresponds to a point
in $\Gamma.$ The weights of the edges are chosen according to a weight
function $w_{\varepsilon}\left(x,y\right)$ which measures the similarities
between every pair of points where the parameter $\varepsilon$ defines
a local neighborhood for each point. The weight function is defined
by a kernel function obeying the following properties:
\begin{description}
\item [{symmetry:}] $\forall x_{i},x_{j}\in\Gamma,\,\, w_{\varepsilon}\left(x_{i},x_{j}\right)=w_{\varepsilon}\left(x_{j},x_{i}\right)$ 
\item [{non-negativity:}] $\forall x_{i},x_{j}\in\Gamma,\,\, w_{\varepsilon}\left(x_{i},x_{j}\right)\ge0$ 
\end{description}
\textbf{positive semi-definite:} for every real-valued bounded function
$f$ defined on $\Gamma$, $\sum_{x_{i},x_{j}\in\Gamma}w_{\varepsilon}\left(x_{i},x_{j}\right)f\left(x_{i}\right)f\left(x_{j}\right)\ge0.$ 
\begin{description}
\item [{fast~decay:}] $w_{\varepsilon}\left(x_{i},x_{j}\right)\rightarrow0$
when $\left\Vert x_{i}-x_{j}\right\Vert \gg\varepsilon$ and $w_{\varepsilon}\left(x_{i},x_{j}\right)\rightarrow1$
when $\left\Vert x_{i}-x_{j}\right\Vert \ll\varepsilon$. This property
facilitates the representation of $w_{\varepsilon}$ by a sparse matrix.
\end{description}
A common choice that meets these criteria is the Gaussian kernel:
\[
w_{\varepsilon}\left(x_{i},x_{j}\right)=e^{-\frac{\left\Vert x_{i}-x_{j}\right\Vert ^{2}}{2\varepsilon}}.
\]

A weight matrix $w_{\varepsilon}$ is used to represent the weights
of the edges. Given a graph $G$, the Graph Laplacian normalization
\citep{C97} is applied to the weight matrix $w_{\varepsilon}$ and
the result is given by $M$: 
\[
M_{i,j}\triangleq m\left(x,y\right)=\frac{w_{\varepsilon}\left(x,y\right)}{d\left(x\right)}
\]
where $d(x)=\sum_{y\in\Gamma}w_{\varepsilon}\left(x,y\right)$ is
the degree of $x$. This transforms $w_{\varepsilon}$ into a Markov
transition matrix corresponding to a random walk through the points
in $\Gamma$. The probability to move from $x$ to $y$ in \emph{one}
time step is denoted by $m\left(x,y\right)$. These probabilities
measure the connectivity of the points within the graph.

The transition matrix $M$ is conjugate to a symmetric matrix $A$
whose elements are given by $A_{i,j}\triangleq a\left(x,y\right)=\sqrt{d\left(x\right)}m\left(x,y\right)\frac{1}{\sqrt{d(y)}}.$
Using matrix notation, $A$ is given by $A=D^{\frac{1}{2}}MD^{-\frac{1}{2}},$
where $D$ is a diagonal matrix whose values are given by $d\left(x\right)$.
The matrix $A$ has $n$ real eigenvalues $\left\{ \lambda_{l}\right\} _{l=0}^{n-1}$
where $0\le\lambda_{l}\le1,$ and a set of orthonormal eigenvectors
$\left\{ v_{l}\right\} _{l=1}^{N-1}$ in $\mathbb{R}^{n}$. Thus,
$A$ has the following spectral decomposition: 
\begin{equation}
a\left(x,y\right)=\sum_{k\ge0}\lambda_{k}v_{l}\left(x\right)v_{l}\left(y\right).\label{A_spectral}
\end{equation}
Since $M$ is conjugate to $A$, the eigenvalues of both matrices
are identical. In addition, if $\left\{ \phi_{l}\right\} $ and $\left\{ \psi_{l}\right\} $
are the left and right eigenvectors of $M$, respectively, then the
following equalities hold: 
\begin{equation}
\phi_{l}=D^{\frac{1}{2}}v_{l},\;\;\;\;\psi_{l}=D^{-\frac{1}{2}}v_{l}.\label{P_eigenvecs}
\end{equation}

From the orthonormality of $\left\{ v_{i}\right\} $ and Eq. \ref{P_eigenvecs}
it follows that $\left\{ \phi_{l}\right\} $ and $\left\{ \psi_{l}\right\} $
are bi-orthonormal i.e. $\langle\phi_{m},\psi_{l}\rangle=\delta_{ml}$
where $\delta_{ml}=1$ when $m=l$ and $\delta_{ml}=0$, otherwise.
Combing Eqs. \ref{A_spectral} and \ref{P_eigenvecs} together with
the bi-orthogonality of $\left\{ \phi_{l}\right\} $ and $\left\{ \psi_{l}\right\} $
leads to the following eigen-decomposition of the transition matrix
$M$ 
\begin{equation}
m\left(x,y\right)=\sum_{l\ge0}\lambda_{l}\psi_{l}\left(x\right)\phi_{l}\left(y\right).\label{P_spectral}
\end{equation}
When the spectrum decays rapidly (provided $\varepsilon$ is appropriately
chosen - see Sec. \ref{app:choosing_epsilon}), only a few terms are
required to achieve a given accuracy in the sum. Namely,
\[
m\left(x,y\right)\backsimeq\sum_{l=0}^{n\left(p\right)}\lambda_{l}\psi_{l}\left(x\right)\phi_{l}\left(y\right)
\]
where $n\left(p\right)$ is the number of terms which are required
to achieve a given precision $p$.

We recall the \emph{diffusion distance }between two data points $x$
and $y$ as it was defined in \citep{CL_DM06}: 
\begin{equation}
D^{2}\left(x,y\right)=\sum_{z\in\Gamma}\frac{\left(m\left(x,z\right)-m\left(z,y\right)\right)^{2}}{\phi_{0}\left(z\right)}.\label{diffusion_distance}
\end{equation}
This distance reflects the geometry of the dataset and it depends
on the number of paths connecting $x$ and $y$. Substituting Eq.
\ref{P_spectral} in Eq. \ref{diffusion_distance} together with the
bi-orthogonality property allows to express the diffusion distance
using the right eigenvectors of the transition matrix $M$: 
\begin{equation}
D^{2}\left(x,y\right)=\sum_{l\ge1}\lambda_{l}^{2}\left(\psi_{l}\left(x\right)-\psi_{l}\left(y\right)\right)^{2}.\label{diffusion_distance_psi}
\end{equation}
Thus, the family of Diffusion Maps $\left\{ \Psi(x)\right\} $ which
is defined by 
\begin{equation}
\Psi\left(x\right)=\left(\lambda_{1}\psi_{1}\left(x\right),\lambda_{2}\psi_{2}\left(x\right),\lambda_{3}\psi_{3}\left(x\right),\cdots\right)\label{eq:DM_family}
\end{equation}
embeds the dataset into a Euclidean space. In the new coordinates
of Eq. \ref{eq:DM_family}, the \emph{Euclidean }distance between
two points in the embedding space is equal to the \emph{diffusion
}distance between their corresponding two high dimensional points
as defined by the random walk. Moreover, this facilitates the embedding
of the original points into a low-dimensional Euclidean space $\mathbb{R}^{q}$
by: 
\begin{equation}
\Xi_{t}:x_{i}\rightarrow\left(\lambda_{2}^{t}\psi_{2}\left(x_{i}\right),\lambda_{3}^{t}\psi_{3}\left(x_{i}\right),\dots,\lambda_{q+1}^{t}\psi_{q+1}\left(x_{i}\right)\right).\label{eq:DM_embedding}
\end{equation}
which also endows coordinates on the set $\Gamma$. Since $\lambda_{1}=1$
and $\psi_{1}\left(x\right)$ is constant, the embedding uses $\lambda_{2},\ldots,\lambda_{q+1}$.
Essentially, $q\ll n$ due to the fast decay of the eigenvalues of
$M$. Furthermore, $q$ depends only on the dimensionality of the
data as captured by the random walk and not on the original dimensionality
of the data. Diffusion maps have been successfully applied for acoustic
detection of moving vehicles \citep{SchclarDetection2010} and fusion
of data and multicue data matching \citep{Lafon06datafusion}.

\subsection{Choosing $\varepsilon$\label{app:choosing_epsilon} }

The choice of $\varepsilon$ is critical to achieve the optimal performance
by the DM algorithm since it defines the size of the local neighborhood
of each point. On one hand, a large $\varepsilon$ produces a coarse
analysis of the data as the neighborhood of each point will contain
a large number of points. In this case, the diffusion distance will
be close to $1$ for most pairs of points. On the other hand, a small
$\varepsilon$ might produce many neighborhoods that contain only
a single point. In this case, the diffusion distance is zero for most
pairs of points. The best choice lies between these two extremes.
Accordingly, the ensemble classifier which is based on the the Diffusion
Maps algorithm will construct different versions of the training set
using different values of $\varepsilon$ which will be chosen between
the shortest and longest pairwise distances.

\subsection{The Nyström out-of-sample extension \label{sec:OUT-OF-SAMPLE-EXTENSION}}

The Nyström extension \citep{N28} is an extrapolation method that
facilitates the extension of any function $f:\Gamma\rightarrow\mathbb{R}$
to a set of new points which are added to $\Gamma$. Such extensions
are required in on-line processes in which new samples arrive and
a function $f$ that is defined on $\Gamma$ needs to be extrapolated
to include the new points. These settings exactly fit the settings
of the proposed approach since the test samples are given \emph{after}
the dimensionality of the training set was reduced. Specifically,
the Nyström extension is used to embed a new point into the reduced-dimension
space where every coordinate of the low-dimensional embedding constitutes
a function that needs to be extended. 

We describe the Nyström extension scheme for the Gaussian kernel that
is used by the Diffusion Maps algorithm. Let $\Gamma$ be a set of
points in $\mathbb{R}^{n}$ and $\Psi$ be its embedding (Eq. \ref{eq:DM_family}).
Let $\bar{\Gamma}$ be a set in $\mathbb{R}^{n}$ such that $\Gamma\subset\bar{\Gamma}$.
The Nyström extension scheme extends $\Psi$ onto the dataset $\bar{\Gamma}$.
Recall that the eigenvectors and eigenvalues form the dimension-reduced
coordinates of $\Gamma$ (Eq. \ref{eq:DM_embedding}). The eigenvectors
and eigenvalues of a Gaussian kernel with width $\varepsilon$ which
is used to measure the pairwise similarities in the training set $\Gamma$
are computed according to 
\begin{equation}
\lambda_{l}\varphi_{l}\left(x\right)=\sum_{y\in\Gamma}e^{-\frac{\parallel x-y\parallel^{2}}{2\varepsilon}}\varphi_{l}\left(y\right),\mbox{ \ensuremath{x\in\Gamma.}}\label{extension}
\end{equation}
 If $\lambda_{l}\neq0$ for every $l$, the eigenvectors in Eq. \ref{extension}
can be extended to any $x\in\mathbb{R}^{n}$ by 
\begin{equation}
\bar{\varphi}_{l}\left(x\right)=\frac{1}{\lambda_{l}}\sum_{y\in\Gamma}e^{-\frac{\parallel x-y\parallel^{2}}{2\varepsilon}}\varphi_{l}\left(y\right),\mbox{ \ensuremath{x\in\mathbb{R}^{n}}.}\label{nystrom}
\end{equation}

Let $f$ be a function on the training set $\Gamma$ and let $x\notin\Gamma$
be a new point. In the Diffusion Maps setting, we are interested in
approximating 
\[
\Psi\left(x\right)=\left(\lambda_{2}\psi_{2}\left(x\right),\lambda_{3}\psi_{3}\left(x\right),\cdots,\lambda_{q+1}\psi_{q+1}\left(x\right)\right).
\]
The eigenfunctions $\left\{ \varphi_{l}\right\} $ are the outcome
of the spectral decomposition of a symmetric positive matrix. Thus,
they form an orthonormal basis in $\mathbb{R}^{N}$ where $N$ is
the number of points in $\Gamma$. Consequently, any function $f$
can be written as a linear combination of this basis: 
\[
f\left(x\right)=\sum_{l}\langle\varphi_{l},f\rangle\varphi_{l}\left(x\right),\mbox{ \ensuremath{x\in\Gamma.}}
\]
Using the Nyström extension, as given in Eq. \ref{nystrom}, $f$
can be defined for any point in $\mathbb{R}^{n}$ by

\begin{equation}
\bar{f}\left(x\right)=\sum_{l}\langle\varphi_{l},f\rangle\bar{\varphi_{l}}\left(x\right),\mbox{ \ensuremath{x\in\mathbb{R}^{n}.}}\label{f_extend}
\end{equation}

The above extension facilitates the decomposition of every diffusion
coordinate $\psi_{i}$ as $\psi_{i}(x)=\sum_{l}\langle\varphi_{l},\psi_{i}\rangle\varphi_{l}\left(x\right),\mbox{ \ensuremath{x\in\Gamma}}$.
In addition, the embedding of a new point $\bar{x}\in\bar{\Gamma}\backslash\Gamma$
can be evaluated in the embedding coordinate system by $\bar{\psi_{i}}\left(\bar{x}\right)=\sum_{l}\langle\varphi_{l},\psi_{i}\rangle\bar{\varphi_{l}}\left(\bar{x}\right)$. 

Note that the scheme is ill conditioned since $\lambda_{l}\longrightarrow0$
as $l\longrightarrow\infty$. This can be solved by cutting-off the
sum in Eq. \ref{f_extend} and keeping only the eigenvalues (and their
corresponding eigenfunctions) that satisfy $\lambda_{l}\ge\delta\lambda_{0}$
(where $0<\delta\leq1$ and the eigenvalues are given in descending
order of magnitude): 
\begin{equation}
\bar{f}\left(x\right)=\sum_{\lambda_{l}\ge\delta\lambda_{0}}\langle\varphi_{l},f\rangle\bar{\varphi_{l}}\left(x\right),\mbox{ \ensuremath{x\in\mathbb{R}^{n}}.}\label{f_extend_delta}
\end{equation}

The result is an extension scheme with a condition number $\delta$.
In this new scheme, $f$ and $\bar{f}$ do not coincide on $\Gamma$
but they are relatively close. The value of $\varepsilon$ controls
this error. Thus, choosing $\varepsilon$ carefully may improve the
accuracy of the extension.

\subsection{Ensemble via Diffusion maps\label{sub:Ensemble-via-Diffusion}}

Let $\Gamma$ be a training set as described in Eq. \ref{eq:training_set}.
Every dimension-reduced version of $\Gamma$ is constructed by applying
the Diffusion Maps algorithm to $\Gamma$ where the parameter $\varepsilon$
is randomly chosen from the set of all pairwise Euclidean distances
between the points in $\Gamma$ i.e. from $\left\{ \parallel x-y\parallel\right\} _{x,y\in\Gamma}$.
The dimension of the reduced space is fixed for all the ensemble members
at a given percentage of the ambient space dimension. We denote by
$\widetilde{\Gamma}\left(\varepsilon_{i}\right)\subseteq\mathbb{R}^{q}$
the training set that is obtained from the application of the diffusion
maps algorithm to $\Gamma$ using the randomly chosen value $\varepsilon_{i}$
where $i=1,\ldots,K$ and $K$ is the number of ensemble members.
The ensemble members are constructed by applying a given induction
algorithm to each training set $\widetilde{\Gamma}\left(\varepsilon_{i}\right)$.
In order to classify a new sample, it is first embedded into the dimension-reduced
space $\mathbb{R}^{q}$ of each classifier using the Nyström extension
(Section \ref{sec:OUT-OF-SAMPLE-EXTENSION}). Then, every ensemble
member classifies the new sample and the voting scheme which is described
in Section \ref{sec:The-proposed-approach} is used to produce the
ensemble classification. Note that in order for the Nyström extension
to work, each ensemble member must store the eigenvectors and eigenvalues
which were produced by the Diffusion Maps algorithm.

\section{Random Projections\label{sec:Random-Projections}}

The Random projections algorithm implements the Johnson and Lindenstrauss
lemma \citep{JL84}\emph{ }(see Section \ref{sub:Dimensionality-reduction})\emph{.
}In order to reduce the dimensionality of a given training set $\Gamma$,
a set of random vectors $\Upsilon=\left\{ \rho_{i}\right\} _{i=1}^{n}$
is generated where $\rho_{i}\in\mathbb{R}^{q}$ are column vectors\emph{
}and \emph{$\left\Vert \rho_{i}\right\Vert _{l_{2}}=1$. }Two common
ways to choose the entries of the vectors $\left\{ \rho_{i}\right\} _{i=1}^{n}$
are:
\begin{enumerate}
\item From a uniform (or normal) distribution over the $q$ dimensional
unit sphere. 
\item From a Bernoulli +1/-1 distribution. In this case, the vectors are
normalized so that $\left\Vert \rho_{i}\right\Vert _{l_{2}}=1$ for
$i=1,\ldots,n$. 
\end{enumerate}
Next, the vectors in $\Upsilon$ are used to form the columns of a
$q\times n$ matrix 
\begin{equation}
R=\left(\rho_{1}|\rho_{2}|\ldots|\rho_{n}\right).\label{eq:Random_proj_matrix}
\end{equation}
The embedding $\widetilde{x}_{i}$ of $x_{i}$ is obtained by \emph{
\[
\widetilde{x}_{i}=R\cdot x_{i}
\]
}

Random projections are well suited for the construction of ensembles
of classifiers since the randomization meets the diversity criterion
(Section \ref{sub:Ensemble-Classifiers}) while the bounded distortion
rate provides the accuracy. 

Random projections have been successfully employed for dimensionality
reduction in \citep{fern2003random} as part of an ensemble algorithm
for clustering. An Expectation Maximization (of Gaussian mixtures)
clustering algorithm was applied to the dimension-reduced data. %
{} The ensemble algorithm achieved results that were superior to those
obtained by: (a) a single run of random projection/clustering; and
(b) a similar scheme which used PCA to reduce the dimensionality of
the data.

\subsection{Out-of-sample extension\label{sub:Out-of-sample-extension}}

In order to embed a new sample $y$ into the dimension-reduced space
$\mathbb{R}^{q}$ of the \emph{i}-th ensemble member, the sample is
simply projected onto the random matrix $R$ that was used to reduce
the dimensionality of the member's training set. The embedding of
$y$ is given by $\tilde{y}=R\cdot y$. Accordingly, each random matrix
needs to be stored as part of its corresponding ensemble member in
order to allow out-of-sample extension.

\subsection{Ensemble via Random Projections}

In order to construct the dimension-reduced versions of the training
set, $K$ random matrices $\left\{ R_{i}\right\} _{i=1}^{K}$ are
constructed (recall that $K$ is the number of ensemble members).
The training set is projected onto each random matrix $R_{i}$ and
the dataset which is produced by each projection is denoted by $\Gamma\left(R_{i}\right)$.
The ensemble members are constructed by applying a given inducer to
each of the dimension-reduced datasets in $\left\{ \Gamma\left(R_{i}\right)\right\} _{i=1}^{K}$. 

A new sample is classified by first embedding it into the dimension-reduced
space $\mathbb{R}^{q}$ of every classifier using the scheme in Section
\ref{sub:Out-of-sample-extension}. Then, each ensemble member classifies
the new sample and the voting scheme from Section \ref{sec:The-proposed-approach}
is used to determine the classification by the ensemble.

\section{Random Subspaces\label{sec:Random-Subspaces}}

The Random subspaces algorithm reduces the dimensionality of a given
training set $\Gamma$ by projecting the vectors onto a random subset
of attributes. Formally, let $\left\{ i_{k}\right\} _{k=1}^{q}$ be
a randomly chosen subset of attributes. The embedding $\widetilde{x}$
of $x=\left(x_{1},\ldots,x_{n}\right)$ is obtained by \emph{$\widetilde{x}=\left(x_{i_{1}},\ldots,x_{i_{q}}\right)$}.
Accordingly, each random set of attributes needs to be stored as part
of its corresponding ensemble member.

This method is a special case of the random projections dimensionality
reduction algorithm described in Sec. \ref{sec:Random-Projections}
where the rows (and column) of the matrix $R$ in eq. \ref{eq:Random_proj_matrix}
are unique indicator vectors. 

Random subspaces have been used to construct decision forests \citep{RandomSubspaceDecisionForest}
- an ensemble of tree classifiers - and also to construct ensemble
\emph{regressors }\citep{RandSubSpaceEnsemble04}\emph{.} Ensemble
regressors employ a multivariate function instead of a voting scheme
to combine the individual results of the ensemble members. The training
sets that are constructed by the Random subspaces method are dimension-reduced
versions of the original dataset and therefore this method is investigated
in our experiments. This method combined with support vector machines
has been successfully applied to relevance feedback in image retrieval
\citep{Asymmetric_Bagging_Image_Retrieval}.

\subsection{Out-of-sample extension\label{sub:Out-of-sample-extension-1}}

In order to embed a new sample $y$ into the dimension-reduced space
$\mathbb{R}^{q}$ of the \emph{i}-th ensemble member, the sample is
simply projected onto $\left\{ i_{k}\right\} _{k=1}^{q}$ - the member's
subset of attributes. The embedding of $y=\left(y,\ldots,y_{n}\right)$
is given by $\tilde{y}=\left(y_{i_{1}},\ldots,y_{i_{q}}\right)$.

\subsection{Ensemble via Random Subspaces}

In order to construct the dimension-reduced versions of the training
set, $K$ subsets of features are randomly chosen. The training set
is projected onto each attribute subset and the ensemble members are
constructed by applying a given inducer to each of the dimension-reduced
datasets. 

A new sample is classified by first embedding it into the dimension-reduced
space $\mathbb{R}^{q}$ of every classifier using the scheme in Section
\ref{sub:Out-of-sample-extension-1}. Then, each ensemble member classifies
the new sample and the voting scheme from Section \ref{sec:The-proposed-approach}
is used to determine the ensemble's classification.

\section{Experimental results \label{sec:Experimental-results}}

In order to evaluate the proposed approach, we used the WEKA framework
\citep{WEKA}. We tested our approach on $17$ datasets from the UCI
repository \citep{UCI} which contains benchmark datasets that are
commonly used to evaluate machine learning algorithms. The list of
datasets and their properties are summarized in Table \ref{tab:datasets-1}.

\begin{table*}[t]
\begin{centering}
\caption{Properties of the benchmark datasets used for the evaluation. \label{tab:datasets-1}}

\par\end{centering}

\centering{}%
\begin{tabular}{|c|c|c|c|}
\hline 
\textbf{Dataset Name}  & \textbf{Instances}  & \textbf{Features}  & \textbf{Labels}\tabularnewline
\hline 
Musk1  & 476 & 166 & 2\tabularnewline
\hline 
Musk2  & 6598 & 166 & 2\tabularnewline
\hline 
Pima-diabetes & 768 & 8 & 2\tabularnewline
\hline 
Ecoli  & 335 & 7 & 8\tabularnewline
\hline 
Glass  & 214 & 9 & 7\tabularnewline
\hline 
Hill Valley with noise & 1212 & 100 & 2\tabularnewline
\hline 
Hill Valley without noise & 1212 & 100 & 2\tabularnewline
\hline 
Ionosphere  & 351  & 34 & 2\tabularnewline
\hline 
Iris & 150 & 4 & 3\tabularnewline
\hline 
Isolet & 7797 & 617 & 26\tabularnewline
\hline 
Letter & 20000 & 16 & 26\tabularnewline
\hline 
Madelon & 2000 & 500  & 2\tabularnewline
\hline 
Multiple features  & 2000 & 649  & 10\tabularnewline
\hline 
Sat & 6435 & 36 & 7\tabularnewline
\hline 
Waveform with noise  & 5000 & 40 & 3\tabularnewline
\hline 
Waveform without noise  & 5000 & 21 & 3\tabularnewline
\hline 
Yeast & 1484 & 8 & 10\tabularnewline
\hline 
\end{tabular}
\end{table*}

{}

\subsection{Experiment configuration}

In order to reduce the dimensionality of a given training set, one
of two schemes was employed depending on the dimensionality reduction
algorithm at hand. The first scheme was used for the Random Projection
and the Random Subspaces algorithms and it applied the dimensionality
reduction algorithm to the dataset without any pre-processing of the
dataset. However, due to the space and time complexity of the Diffusion
Maps algorithm, which is quadratic in the size of the dataset, a different
scheme was used. First, a random value $\varepsilon\in\left\{ \parallel x-y\parallel\right\} _{x,y\in\Gamma}$
was selected. Next, a random sample of $600$ unique data items was
drawn (this size was set according to time and memory limitations).
The Diffusion Maps algorithm was then applied to the sample which
produced a dimension-reduced training set. This set was then extended
using the Nyström extension to include the training samples which
were not part of the sample. These steps are summarized in Algorithm
\ref{alg:Steps-for-DM}.

\begin{algorithm}
\textbf{Input:} Dataset $\Gamma$, target dimension $q$

\textbf{Output:} A dimension reduced training set $\widetilde{\widetilde{\Gamma}}$.
\begin{enumerate}
\item Select a random value $\varepsilon\in\left\{ \parallel x-y\parallel\right\} _{x,y\in\Gamma}$ 
\item Select a random sample $\bar{\Gamma}$ of $600$ unique elements from
$\Gamma$.
\item Apply the Diffusion Maps algorithm to $\bar{\Gamma}$ resulting in
$\widetilde{\Gamma}$
\item Extend $\widetilde{\Gamma}$ to include the points in $\Gamma\backslash\bar{\Gamma}$
using the Nyström extension - resulting in $\widetilde{\widetilde{\Gamma}}$.
\end{enumerate}
\caption{Steps for constructing the training set of a single ensemble member
using the Diffusion Maps algorithm. \label{alg:Steps-for-DM}}
\end{algorithm}

All ensemble algorithms were tested using the following inducers:
(a) nearest-neighbors (WEKA's B1 inducer); (b) decision tree (WEKA's
J48 inducer); and (c) Naïve Bayes. The ensembles were composed of
ten classifiers\uline{ }(the information theoretic problem of choosing
the optimal size of an ensemble is out of the scope of this paper.
This problem is discussed, for example, in \citep{Kuncheva:2004})\uline{.}
The dimension-reduced space was set to half of the original dimension
of the data. Ten-fold cross validation was used to evaluate each ensemble's
performance on each of the datasets. 

The constructed ensemble classifiers were compared with: a non-ensemble
classifier which applied the induction algorithm to the dataset without
dimensionality reduction (we refer to this classifier as the \emph{plain
}classifier). The constructed ensemble classifiers were also compared
with the Bagging \citep{Bagging96}, AdaBoost \citep{AdaBoost96}
and Rotation Forest \citep{RotationForest2006} ensemble algorithms.
In order to see whether the Diffusion Maps ensemble classifier can
be further improved as part of a multi-strategy ensemble (Section
\ref{sub:Ensemble-Classifiers}), we constructed an ensemble classifier
whose members applied the AdaBoost algorithm to their Diffusion Maps
dimension-reduced training sets. 

We used the default values of the parameters of the WEKA built-in
ensemble classifiers in all the experiments. For the sake of simplicity,
in the following we refer to the ensemble classifiers which use the
Diffusion Maps and Random Projections dimensionality algorithms as
the DME and RPE classifiers, respectively. The ensemble classifier
which is based on the random subspaces dimensionality reduction algorithm
is referred to as the RSE classifier.

\subsection{Results}

Tables \ref{tab:TestIB1Full}, \ref{tab:TestJ48Full}, and \ref{tab:TestBayesFull}
describe the results obtained by the decision tree, nearest-neighbor
and Naïve Bayes inducers, respectively. In each of the tables, the
first column specifies the name of the tested dataset and the second
column contains the results of the plain classifier. The second to
last row contains the average improvement percentage of each algorithm
compared to the plain classifier. We calculate the average rank of
each inducer across all datasets in the following manner: for each
of the datasets, the algorithms are ranked according to the accuracy
that they achieved. The average rank of a given inducer is obtained
by averaging its obtained ranks over all the datasets. The average
rank is given in the last row of each table. 

The results of the experimental study indicate that dimensionality
reduction is a promising approach for the construction of ensembles
of classifiers. In 113 out of 204 cases the dimensionality reduction
ensembles outperformed the plain algorithm with the following distribution:
RPE (33 cases out of 113), DM+AdaBoost (30 cases), RSE (27 cases)
and DM (23 cases). 

\begin{figure}
\centering{}\includegraphics[bb=55bp 120bp 787bp 481bp,clip,width=0.9\columnwidth]{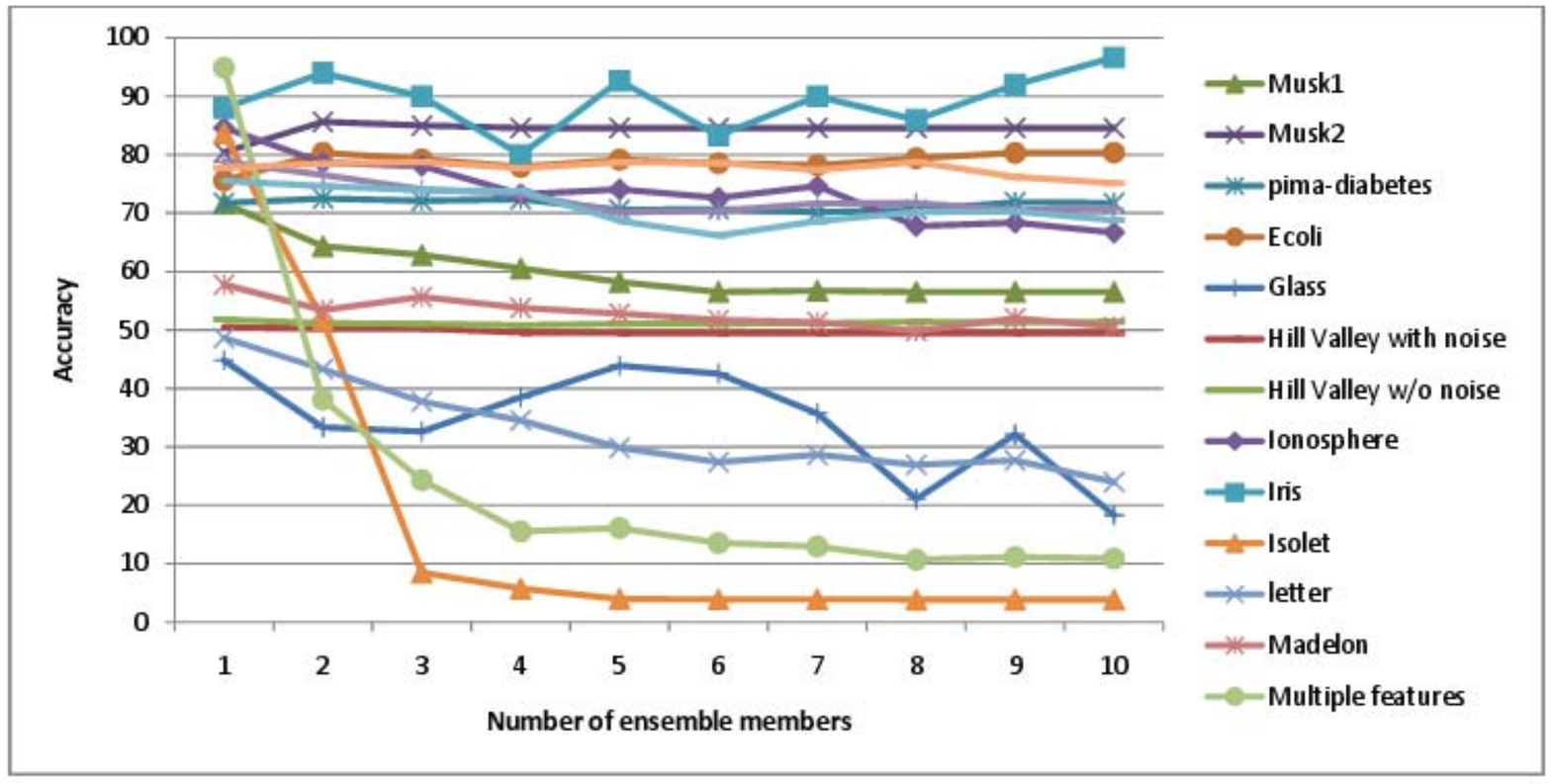}\caption{Accuracy of the RSE algorithm using the Naïve Bayes inducer.\label{fig:RSE-NB-accuracy}}
\end{figure}

\begin{figure}
\centering{}\includegraphics[bb=55bp 120bp 785bp 480bp,clip,width=0.9\columnwidth]{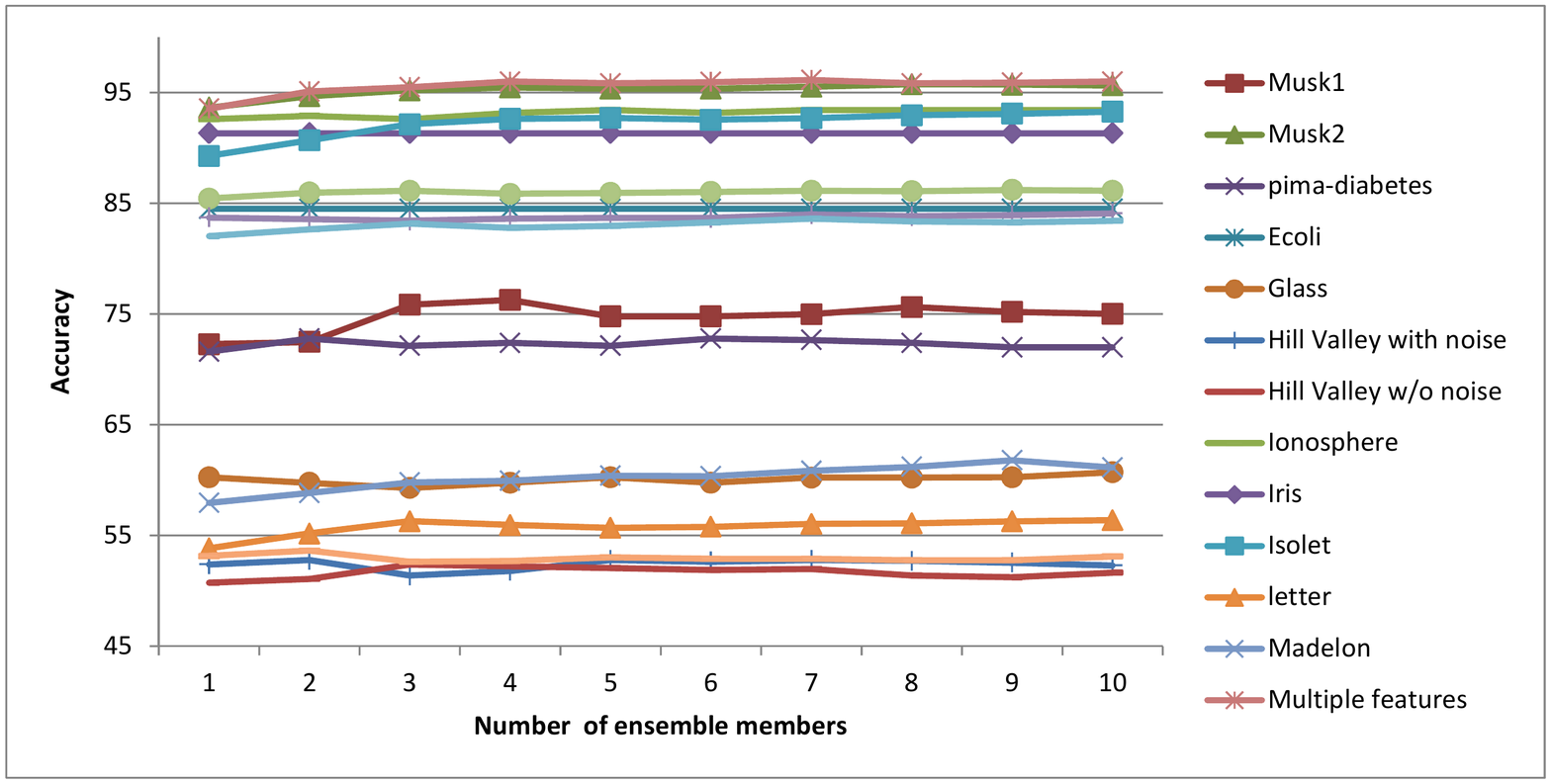}\caption{Accuracy of the DM ensemble using the Naïve Bayes inducer.\label{fig:DM-NB-accuracy}}
\end{figure}

Ranking all the algorithms according to the average accuracy improvement
percentage produces the following order: Rotation Forest (6.4\%),
Random projection (4\%), DM+AdaBoost (2.1\%), Bagging (1.5\%), AdaBoost
(1\%), DM (0.7\%) and Random subspaces (-6.7\%). Note that the RSE
algorithm achieved an average decrease of 6.7\% in accuracy. A closer
look reveals that this was caused by a particularly bad performance
when the Naïve Bayes inducer was used (26\% average decrease in accuracy).
In contrast, improvement averages of 1.7\% and 4.4\% were achieved
when the RSE algorithm used the nearest-neighbors and J48 inducers,
respectively. This may be due to datasets whose features are not independent
- a situation which does not conform with the basic assumption of
the Naïve Bayes inducer. For example, the \emph{Isolet} dataset is
composed of acoustic recordings that are decomposed to \emph{overlapping}
segments where features of each segment constitute an instance in
the dataset. In these settings, the features are not independent.
Since the other algorithms, including the plain one, achieve much
better results when applied to this dataset, we can assume that because
the RSE algorithm chooses a random subset of features, the chance
of obtaining independent features is lower compared to when all features
are selected. Moreover, given the voting scheme in Section \ref{sec:The-proposed-approach},
ensemble members which produce wrong classifications with high probabilities
damage accurate classifications obtained by other ensemble members.
Figure \ref{fig:RSE-NB-accuracy} demonstrates how the accuracy decreases
as the number of members increases when RSE is paired with the Naïve
Bayes inducer. This phenomenon is contrasted in Fig. \ref{fig:DM-NB-accuracy}
where the behavior that is expected from the ensemble is observed.
Namely, an increase in accuracy when the number of ensemble members
is increased when an ensemble different from the RSE is used (e.g.
the DME).

In order to compare the 8 algorithms across all inducers and datasets
we applied the procedure presented in \citep{demsar2006}. The null
hypothesis that all methods have the same accuracy could not be rejected
by the adjusted Friedman test with a confidence level of 90\% (specifically
F(7,350)=0.79 < 1.73 with p-value>0.1). Furthermore, the results show
there is a dependence between the inducer, dataset and chosen dimensionality
reduction algorithm. In the following we investigate the dependence
between the latter two for each of the inducers.

\subsubsection{Results for the nearest neighbor inducer (IB1)}

In terms of the average improvement, the RPE algorithm is ranked first
with an average improvement percentage of 5.8\%. We compared the various
algorithms according to their average rank following the steps described
in \citep{demsar2006}. The RSE and RPE achieved the first and second
average rank, respectively. They were followed by Bagging ($3^{rd}$)
and Rotation Forest ($4^{th}$).

Using the adjusted Friedman test we rejected the null hypothesis that
all methods achieve the same classification accuracy with a confidence
level of 95\% and (7, 112) degrees of freedom (specifically F(7, 112)=2.47
> 2.09 and p-value<0.022). Following the rejection of the null hypothesis,
we employed the Nemenyi post-hoc test where in the experiment settings
two classifiers are significantly different with a confidence level
of 95\% if their average ranks differ by at least $CD=2.55$. The
null hypothesis that any of the non-plain algorithms has the same
accuracy as the plain algorithm could not be rejected at confidence
level 95\%.

\begin{sidewaystable}
\caption{\label{tab:TestIB1Full}Results of the ensemble classifiers based
on the nearest-neighbor inducer (WEKA's IB1).}

\textbf{RPE }is the Random Projection ensemble algorithm; \textbf{RSE}
is the Random Subspaces ensemble algorithm; \textbf{DME} is the Diffusion
Maps ensemble classifier; \textbf{DME+AdaBoost} is the multi-strategy
ensemble classifier which applied AdaBoost to the Diffusion Maps dimension-reduced
datasets.

\centering{}{\scriptsize }%
\begin{tabular}{lcccccccc}
 &  &  &  &  &  &  &  & \tabularnewline
\hline 
\textbf{\scriptsize Dataset } & \multicolumn{1}{c}{\textbf{\scriptsize Plain NN}} & \multicolumn{1}{c}{\textbf{\scriptsize RPE}} & \multicolumn{1}{c}{\textbf{\scriptsize RSE}} & \multicolumn{1}{c}{\textbf{\scriptsize Bagging}} & \multicolumn{1}{c}{\textbf{\scriptsize DME}} & \multicolumn{1}{c}{\textbf{\scriptsize DME+AdaBoost}} & \multicolumn{1}{c}{\textbf{\scriptsize AdaBoost}} & \multicolumn{1}{c}{\textbf{\scriptsize Rotation Forest}}\tabularnewline
\hline 
\textbf{\scriptsize Musk1} & {\scriptsize 84.89 $\pm$ 4.56} & {\scriptsize 86.15 $\pm$ 2.94 } & {\scriptsize 86.98 $\pm$ 4.18} & {\scriptsize 86.77 $\pm$ 4.32 } & {\scriptsize 84.46 $\pm$ 4.31 } & {\scriptsize 84.87 $\pm$ 4.52} & {\scriptsize 87.42 $\pm$ 4.24 } & {\scriptsize 84.88 $\pm$ 3.92}\tabularnewline
\textbf{\scriptsize Musk2 } & {\scriptsize 95.80 $\pm$ 0.34 } & {\scriptsize 95.62 $\pm$ 0.38} & {\scriptsize 96.04 $\pm$ 0.33} & {\scriptsize 95.89 $\pm$ 0.31} & {\scriptsize 95.39 $\pm$ 0.39} & {\scriptsize 95.94 $\pm$ 0.49} & {\scriptsize 96.03 $\pm$ 0.35} & {\scriptsize 95.60 $\pm$ 0.62}\tabularnewline
\textbf{\scriptsize pima-diabetes } & {\scriptsize 70.17 $\pm$ 4.69 } & {\scriptsize 72.14 $\pm$ 4.03 } & {\scriptsize 70.83 $\pm$ 3.58} & {\scriptsize 70.44 $\pm$ 3.89} & {\scriptsize 66.79 $\pm$ 4.58} & {\scriptsize 66.40 $\pm$ 4.82} & {\scriptsize 67.30 $\pm$ 5.61} & {\scriptsize 70.04 $\pm$ 4.17}\tabularnewline
\textbf{\scriptsize Ecoli } & {\scriptsize 80.37 $\pm$ 6.38 } & {\scriptsize 83.02 $\pm$ 3.52} & {\scriptsize 83.05 $\pm$ 6.94} & {\scriptsize 80.96 $\pm$ 5.43} & {\scriptsize 77.37 $\pm$ 6.63} & {\scriptsize 76.48 $\pm$ 8.23} & {\scriptsize 78.87 $\pm$ 7.19} & {\scriptsize 81.56 $\pm$ 4.97}\tabularnewline
\textbf{\scriptsize Glass} & {\scriptsize 70.52 $\pm$ 8.94 } & {\scriptsize 76.67 $\pm$ 7.22} & {\scriptsize 77.58 $\pm$ 6.55} & {\scriptsize 70.52 $\pm$ 8.94} & {\scriptsize 72.88 $\pm$ 8.51} & {\scriptsize 71.97 $\pm$ 7.25} & {\scriptsize 70.95 $\pm$ 8.12} & {\scriptsize 70.04 $\pm$ 8.24}\tabularnewline
\textbf{\scriptsize Hill Valley with noise} & {\scriptsize 59.83 $\pm$ 5.48 } & {\scriptsize 68.74 $\pm$ 3.58} & {\scriptsize 59.75 $\pm$ 4.29} & {\scriptsize 59.74 $\pm$ 4.77} & {\scriptsize 50.49 $\pm$ 4.75} & {\scriptsize 50.41 $\pm$ 4.49} & {\scriptsize 58.42 $\pm$ 3.80} & {\scriptsize 79.30 $\pm$ 3.60}\tabularnewline
\textbf{\scriptsize Hill Valley w/o noise} & {\scriptsize 65.84 $\pm$ 4.31 } & {\scriptsize 79.21 $\pm$ 3.19} & {\scriptsize 66.66 $\pm$ 4.48} & {\scriptsize 65.67 $\pm$ 4.26} & {\scriptsize 55.36 $\pm$ 5.60} & {\scriptsize 54.45 $\pm$ 5.18} & {\scriptsize 63.20 $\pm$ 4.28} & {\scriptsize 92.74 $\pm$ 2.10}\tabularnewline
\textbf{\scriptsize Ionosphere} & {\scriptsize 86.33 $\pm$ 4.59 } & {\scriptsize 90.02 $\pm$ 5.60} & {\scriptsize 90.30 $\pm$ 4.32} & {\scriptsize 86.90 $\pm$ 4.85} & {\scriptsize 92.88 $\pm$ 4.09} & {\scriptsize 93.44 $\pm$ 4.68} & {\scriptsize 87.48 $\pm$ 3.55} & {\scriptsize 86.61 $\pm$ 4.26}\tabularnewline
\textbf{\scriptsize Iris } & {\scriptsize 95.33 $\pm$ 5.49 } & {\scriptsize 93.33 $\pm$ 8.31} & {\scriptsize 92.00 $\pm$ 10.80} & {\scriptsize 96.00 $\pm$ 4.66} & {\scriptsize 94.00 $\pm$ 5.84} & {\scriptsize 94.00 $\pm$ 5.84} & {\scriptsize 95.33 $\pm$ 5.49} & {\scriptsize 94.00 $\pm$ 5.84}\tabularnewline
\textbf{\scriptsize Isolet } & {\scriptsize 89.94 $\pm$ 0.71 } & {\scriptsize 90.61 $\pm$ 0.86} & {\scriptsize 90.57 $\pm$ 0.70} & {\scriptsize 89.59 $\pm$ 0.65} & {\scriptsize 91.32 $\pm$ 0.72} & {\scriptsize 91.54 $\pm$ 0.87} & {\scriptsize 89.00 $\pm$ 0.86} & {\scriptsize 89.78 $\pm$ 0.78}\tabularnewline
\textbf{\scriptsize Letter } & {\scriptsize 96.00 $\pm$ 0.60 } & {\scriptsize 93.64 $\pm$ 0.32} & {\scriptsize 94.08 $\pm$ 0.76} & {\scriptsize 96.00 $\pm$ 0.57} & {\scriptsize 90.58 $\pm$ 0.70} & {\scriptsize 90.50 $\pm$ 0.76} & {\scriptsize 95.10 $\pm$ 0.43} & {\scriptsize 96.25 $\pm$ 0.55}\tabularnewline
\textbf{\scriptsize Madelon} & {\scriptsize 54.15 $\pm$ 4.28 } & {\scriptsize 68.95 $\pm$ 3.59} & {\scriptsize 55.65 $\pm$ 2.63} & {\scriptsize 54.80 $\pm$ 3.29} & {\scriptsize 65.60 $\pm$ 1.94} & {\scriptsize 65.10 $\pm$ 2.38} & {\scriptsize 54.35 $\pm$ 4.76} & {\scriptsize 55.20 $\pm$ 3.54}\tabularnewline
\textbf{\scriptsize Multiple features} & {\scriptsize 97.80 $\pm$ 0.63 } & {\scriptsize 95.65 $\pm$ 1.20} & {\scriptsize 97.90 $\pm$ 0.66} & {\scriptsize 97.85 $\pm$ 0.75} & {\scriptsize 95.45 $\pm$ 1.42} & {\scriptsize 95.55 $\pm$ 1.12} & {\scriptsize 97.45 $\pm$ 0.64} & {\scriptsize 97.70 $\pm$ 0.59}\tabularnewline
\textbf{\scriptsize Sat} & {\scriptsize 90.21 $\pm$ 1.16 } & {\scriptsize 91.34 $\pm$ 0.75} & {\scriptsize 91.47 $\pm$ 0.71} & {\scriptsize 90.37 $\pm$ 1.13} & {\scriptsize 89.74 $\pm$ 0.57} & {\scriptsize 89.40 $\pm$ 0.53} & {\scriptsize 89.01 $\pm$ 1.32} & {\scriptsize 90.82 $\pm$ 1.07}\tabularnewline
\textbf{\scriptsize Waveform with noise} & {\scriptsize 73.62 $\pm$ 1.27 } & {\scriptsize 80.14 $\pm$ 1.65} & {\scriptsize 78.14 $\pm$ 2.35} & {\scriptsize 73.74 $\pm$ 1.69} & {\scriptsize 81.78 $\pm$ 0.93} & {\scriptsize 80.72 $\pm$ 0.98} & {\scriptsize 70.80 $\pm$ 2.03} & {\scriptsize 73.94 $\pm$ 1.69}\tabularnewline
\textbf{\scriptsize Waveform w/o noise} & {\scriptsize 76.90 $\pm$ 2.01 } & {\scriptsize 81.22 $\pm$ 0.90} & {\scriptsize 81.22 $\pm$ 1.47} & {\scriptsize 77.14 $\pm$ 1.55} & {\scriptsize 83.92 $\pm$ 1.38} & {\scriptsize 83.12 $\pm$ 1.16} & {\scriptsize 75.08 $\pm$ 1.70} & {\scriptsize 77.72 $\pm$ 1.39}\tabularnewline
\textbf{\scriptsize Yeast} & {\scriptsize 52.29 $\pm$ 2.39 } & {\scriptsize 55.53 $\pm$ 4.39} & {\scriptsize 49.32 $\pm$ 4.44} & {\scriptsize 52.49 $\pm$ 2.16} & {\scriptsize 48.99 $\pm$ 3.15} & {\scriptsize 48.59 $\pm$ 4.31} & {\scriptsize 51.35 $\pm$ 1.84} & {\scriptsize 53.30 $\pm$ 2.44}\tabularnewline
\hline 
\textbf{\footnotesize Average improvement} & - & \textbf{\footnotesize 5.8\%} & \textbf{\footnotesize 1.7\% } & \textbf{\footnotesize 0.4\%} & \textbf{\footnotesize -0.2\%} & \textbf{\footnotesize -0.6\%} & \textbf{\footnotesize -1\%} & \textbf{\footnotesize 4.6\%}\tabularnewline
\hline 
\textbf{\footnotesize Average rank } & \textbf{\footnotesize 4.97} & \textbf{\footnotesize 3.26} & \textbf{\footnotesize 3.15} & \textbf{\footnotesize 4.35} & \textbf{\footnotesize 5.18} & \textbf{\footnotesize 5.29} & \textbf{\footnotesize 5.44} & \textbf{\footnotesize 4.35}\tabularnewline
\hline 
\end{tabular}
\end{sidewaystable}

\subsubsection{Results for the decision tree inducer (J48)}

Inspecting the average improvement, the RPE and RSE algorithms are
ranked second and third, respectively, after the Rotation Forest algorithm.
Following the procedure presented by Demsar \citep{demsar2006}, we
compared the various algorithms according to their average rank. The
RSE and DM+AdaBoost achieved the second and third best average rank,
respectively, after the Rotation Forest algorithm. 

The null hypothesis that all methods obtain the same classification
accuracy was rejected by the adjusted Friedman test with a confidence
level of 95\% and (7, 112) degrees of freedom (specifically F(7, 112)=5.17
> 2.09 and p-value<0.0001). As the null hypothesis was rejected, we
employed the Nemenyi post-hoc test $(CD=2.55)$. Only the Rotation
Forest algorithm significantly outperformed the plain and the DM algorithms.
The null hypothesis that the RPE, RSE, DM and DM+AdaBoost algorithms
have the same accuracy as the plain algorithm could not be rejected
at confidence level 90\%. 

\begin{sidewaystable}
\caption{\label{tab:TestJ48Full}Results of the Random Projection ensemble
classifier based on the decision-tree inducer (WEKA's J48).}

\textbf{RPE }is the Random Projection ensemble algorithm; \textbf{RSE}
is the Random Subspaces ensemble algorithm; \textbf{DME} is the Diffusion
Maps ensemble classifier; \textbf{DME+AdaBoost} is the multi-strategy
ensemble classier which applied AdaBoost to the Diffusion Maps dimension-reduced
datasets.

\centering{}{\scriptsize }%
\begin{tabular}{lcccccccc}
 &  &  &  &  &  &  &  & \tabularnewline
\hline 
\textbf{\scriptsize Dataset } & \multicolumn{1}{c}{\textbf{\scriptsize Plain J48}} & \multicolumn{1}{c}{\textbf{\scriptsize RPE}} & \multicolumn{1}{c}{\textbf{\scriptsize RSE}} & \multicolumn{1}{c}{\textbf{\scriptsize Bagging}} & \multicolumn{1}{c}{\textbf{\scriptsize DME}} & \multicolumn{1}{c}{\textbf{\scriptsize DME+AdaBoost}} & \multicolumn{1}{c}{\textbf{\scriptsize AdaBoost}} & \multicolumn{1}{c}{\textbf{\scriptsize Rotation Forest}}\tabularnewline
\hline 
\textbf{\scriptsize Musk1} & {\scriptsize 84.90 $\pm$ 6.61} & {\scriptsize 85.31 $\pm$ 6.25} & {\scriptsize 88.45 $\pm$ 8.20} & {\scriptsize 86.56 $\pm$ 6.93} & {\scriptsize 78.60 $\pm$ 7.78} & {\scriptsize 84.89 $\pm$ 5.44} & {\scriptsize 88.46 $\pm$ 6.38} & {\scriptsize 91.60 $\pm$ 3.10}\tabularnewline
\textbf{\scriptsize Musk2 } & {\scriptsize 96.88 $\pm$ 0.63} & {\scriptsize 96.30 $\pm$ 0.78} & {\scriptsize 98.26 $\pm$ 0.39} & {\scriptsize 97.65 $\pm$ 0.50} & {\scriptsize 96.76 $\pm$ 0.72} & {\scriptsize 97.23 $\pm$ 0.67} & {\scriptsize 98.77 $\pm$ 0.35} & {\scriptsize 98.18 $\pm$ 0.67}\tabularnewline
\textbf{\scriptsize pima-diabetes } & {\scriptsize 73.83 $\pm$ 5.66} & {\scriptsize 73.83 $\pm$ 4.86} & {\scriptsize 73.71 $\pm$ 6.04} & {\scriptsize 75.26 $\pm$ 2.96} & {\scriptsize 72.27 $\pm$ 3.11} & {\scriptsize 72.40 $\pm$ 3.68} & {\scriptsize 72.40 $\pm$ 4.86} & {\scriptsize 76.83 $\pm$ 4.80}\tabularnewline
\textbf{\scriptsize Ecoli } & {\scriptsize 84.23 $\pm$ 7.51} & {\scriptsize 86.00 $\pm$ 6.20} & {\scriptsize 84.49 $\pm$ 7.28} & {\scriptsize 84.79 $\pm$ 6.11} & {\scriptsize 83.02 $\pm$ 4.10} & {\scriptsize 81.27 $\pm$ 5.74} & {\scriptsize 83.04 $\pm$ 7.37} & {\scriptsize 86.60 $\pm$ 4.30}\tabularnewline
\textbf{\scriptsize Glass} & {\scriptsize 65.87 $\pm$ 8.91} & {\scriptsize 72.94 $\pm$ 8.19} & {\scriptsize 76.62 $\pm$ 7.38} & {\scriptsize 75.19 $\pm$ 6.40} & {\scriptsize 65.39 $\pm$ 10.54} & {\scriptsize 68.12 $\pm$ 11.07} & {\scriptsize 79.37 $\pm$ 6.13} & {\scriptsize 74.22 $\pm$ 9.72}\tabularnewline
\textbf{\scriptsize Hill Valley with noise} & {\scriptsize 49.67 $\pm$ 0.17} & {\scriptsize 71.28 $\pm$ 4.69} & {\scriptsize 49.67 $\pm$ 0.17} & {\scriptsize 54.62 $\pm$ 3.84} & {\scriptsize 52.39 $\pm$ 3.56} & {\scriptsize 52.39 $\pm$ 5.03} & {\scriptsize 49.67 $\pm$ 0.17} & {\scriptsize 74.51 $\pm$ 2.59}\tabularnewline
\textbf{\scriptsize Hill Valley w/o noise} & {\scriptsize 50.49 $\pm$ 0.17} & {\scriptsize 86.38 $\pm$ 3.77} & {\scriptsize 50.49 $\pm$ 0.17} & {\scriptsize 50.99 $\pm$ 1.28} & {\scriptsize 51.23 $\pm$ 4.40} & {\scriptsize 52.39 $\pm$ 3.34} & {\scriptsize 50.49 $\pm$ 0.17} & {\scriptsize 83.83 $\pm$ 3.94}\tabularnewline
\textbf{\scriptsize Ionosphere} & {\scriptsize 91.46 $\pm$ 3.27} & {\scriptsize 94.32 $\pm$ 3.51} & {\scriptsize 93.75 $\pm$ 4.39} & {\scriptsize 91.75 $\pm$ 3.89} & {\scriptsize 88.04 $\pm$ 4.80} & {\scriptsize 94.87 $\pm$ 2.62} & {\scriptsize 93.17 $\pm$ 3.57} & {\scriptsize 94.89 $\pm$ 3.45}\tabularnewline
\textbf{\scriptsize Iris } & {\scriptsize 96.00 $\pm$ 5.62} & {\scriptsize 95.33 $\pm$ 6.32} & {\scriptsize 94.67 $\pm$ 4.22} & {\scriptsize 94.67 $\pm$ 6.13} & {\scriptsize 92.00 $\pm$ 8.20} & {\scriptsize 90.67 $\pm$ 9.53} & {\scriptsize 93.33 $\pm$ 7.03} & {\scriptsize 96.00 $\pm$ 4.66}\tabularnewline
\textbf{\scriptsize Isolet } & {\scriptsize 83.97 $\pm$ 1.65} & {\scriptsize 87.37 $\pm$ 1.46} & {\scriptsize 92.45 $\pm$ 1.14} & {\scriptsize 90.46 $\pm$ 1.29} & {\scriptsize 90.10 $\pm$ 0.62} & {\scriptsize 93.86 $\pm$ 0.43} & {\scriptsize 93.39 $\pm$ 0.67} & {\scriptsize 93.75 $\pm$ 0.76}\tabularnewline
\textbf{\scriptsize Letter } & {\scriptsize 87.98 $\pm$ 0.51} & {\scriptsize 88.10 $\pm$ 0.52} & {\scriptsize 93.50 $\pm$ 0.92} & {\scriptsize 92.73 $\pm$ 0.69} & {\scriptsize 89.18 $\pm$ 0.79} & {\scriptsize 91.46 $\pm$ 0.78} & {\scriptsize 95.54 $\pm$ 0.36} & {\scriptsize 95.41 $\pm$ 0.46}\tabularnewline
\textbf{\scriptsize Madelon} & {\scriptsize 70.35 $\pm$ 3.78} & {\scriptsize 59.20 $\pm$ 2.57} & {\scriptsize 76.95 $\pm$ 2.69} & {\scriptsize 65.10 $\pm$ 3.73} & {\scriptsize 76.15 $\pm$ 3.43} & {\scriptsize 72.90 $\pm$ 2.27} & {\scriptsize 66.55 $\pm$ 4.09} & {\scriptsize 68.30 $\pm$ 2.98}\tabularnewline
\textbf{\scriptsize Multiple features} & {\scriptsize 94.75 $\pm$ 1.92} & {\scriptsize 95.35 $\pm$ 1.31} & {\scriptsize 97.35 $\pm$ 0.88} & {\scriptsize 96.95 $\pm$ 1.07} & {\scriptsize 93.25 $\pm$ 1.64} & {\scriptsize 94.90 $\pm$ 1.73} & {\scriptsize 97.60 $\pm$ 1.13} & {\scriptsize 97.95 $\pm$ 1.04}\tabularnewline
\textbf{\scriptsize Sat} & {\scriptsize 85.83 $\pm$ 1.04} & {\scriptsize 90.15 $\pm$ 0.93} & {\scriptsize 91.10 $\pm$ 0.91} & {\scriptsize 90.09 $\pm$ 0.78} & {\scriptsize 91.34 $\pm$ 0.48} & {\scriptsize 91.67 $\pm$ 0.37} & {\scriptsize 90.58 $\pm$ 1.12} & {\scriptsize 90.74 $\pm$ 0.69}\tabularnewline
\textbf{\scriptsize Waveform with noise} & {\scriptsize 75.08 $\pm$ 1.33} & {\scriptsize 81.84 $\pm$ 1.43} & {\scriptsize 82.02 $\pm$ 1.50} & {\scriptsize 81.72 $\pm$ 1.43} & {\scriptsize 86.52 $\pm$ 1.78} & {\scriptsize 86.62 $\pm$ 1.76} & {\scriptsize 80.48 $\pm$ 1.91} & {\scriptsize 83.76 $\pm$ 2.07}\tabularnewline
\textbf{\scriptsize Waveform w/o noise} & {\scriptsize 75.94 $\pm$ 1.36} & {\scriptsize 82.56 $\pm$ 1.56} & {\scriptsize 82.52 $\pm$ 1.67} & {\scriptsize 81.48 $\pm$ 1.27} & {\scriptsize 86.96 $\pm$ 1.49} & {\scriptsize 86.36 $\pm$ 0.94} & {\scriptsize 81.46 $\pm$ 1.83} & {\scriptsize 84.94 $\pm$ 1.47}\tabularnewline
\textbf{\scriptsize Yeast} & {\scriptsize 55.99 $\pm$ 4.77} & {\scriptsize 57.82 $\pm$ 3.28} & {\scriptsize 55.32 $\pm$ 4.06} & {\scriptsize 59.23 $\pm$ 3.25} & {\scriptsize 54.85 $\pm$ 3.94} & {\scriptsize 55.39 $\pm$ 2.94} & {\scriptsize 56.39 $\pm$ 5.08} & {\scriptsize 60.71 $\pm$ 3.82}\tabularnewline
\hline 
\textbf{\footnotesize Average improvement} & \textbf{\footnotesize - } & \textbf{\footnotesize 8.5\%} & \textbf{\footnotesize 4.4\% } & \textbf{\footnotesize 3.8\%} & \textbf{\footnotesize 2.2\%} & \textbf{\footnotesize 3.5\%} & \textbf{\footnotesize 3.6\%} & \textbf{\footnotesize 12.2\%}\tabularnewline
\hline 
\textbf{\footnotesize Average rank} & \textbf{\footnotesize 6.26} & \textbf{\footnotesize 4.56} & \textbf{\footnotesize 4.03 } & \textbf{\footnotesize 4.44} & \textbf{\footnotesize 5.68} & \textbf{\footnotesize 4.41} & \textbf{\footnotesize 4.5} & \textbf{\footnotesize 2.15}\tabularnewline
\hline 
\end{tabular}
\end{sidewaystable}

\subsubsection{Results for the Naïve Bayes inducer}

The DM+AdaBoost algorithm achieved the best average improvement and
it is followed by the Rotation Forest algorithm. The DM, RPE and RSE
are ranked $5^{th}$, $7^{th}$ and $8^{th}$ in terms of the average
improvement (possible reasons for the RSE algorithm's low ranking
were described in the beginning of this section). 

Employing the procedure presented in \citep{demsar2006}, we compared
the algorithms according to their average ranks. The DM+AdaBoost and
DM ensembles achieved the second and fourth best average ranks, respectively
while the Rotation Forest and Bagging algorithms achieved the first
and third places, respectively. The null hypothesis that all methods
have the same classification accuracy was rejected by the adjusted
Friedman test with a confidence level of 95\% and (7, 112) degrees
of freedom (specifically F(7, 112)=7.37 > 2.09 and p-value<1e-6).
Since the null hypothesis was rejected, we employed the Nemenyi post-hoc
test. As expected, the RSE was significantly inferior to all other
algorithms. Furthermore, the Rotation Forest algorithm was significantly
better than the RPE algorithms. However, we could not reject at confidence
level 95\% the null hypothesis that the RPE, DM, DM+AdaBoost and the
plain algorithm have the same accuracy. 

\begin{sidewaystable}
\caption{\label{tab:TestBayesFull}Results of the ensemble classifiers based
on the Naïve Bayes inducer.}

\textbf{RPE }is the Random Projection ensemble algorithm; \textbf{RSE}
is the Random Subspaces ensemble algorithm; \textbf{DME} is the Diffusion
Maps ensemble classifier; \textbf{DME+AdaBoost} is the multi-strategy
ensemble classier which applied AdaBoost to the Diffusion Maps dimension-reduced
datasets.

\centering{}{\scriptsize }%
\begin{tabular}{lcccccccc}
 &  &  &  &  &  &  &  & \tabularnewline
\hline 
\textbf{\scriptsize Dataset } & \multicolumn{1}{c}{\textbf{\scriptsize Plain NB }} & \multicolumn{1}{c}{\textbf{\scriptsize RPE}} & \multicolumn{1}{c}{\textbf{\scriptsize RSE}} & \multicolumn{1}{c}{\textbf{\scriptsize Bagging}} & \multicolumn{1}{c}{\textbf{\scriptsize DME}} & \multicolumn{1}{c}{\textbf{\scriptsize DME+AdaBoost}} & \multicolumn{1}{c}{\textbf{\scriptsize AdaBoost}} & \multicolumn{1}{c}{\textbf{\scriptsize Rotation Forest}}\tabularnewline
\hline 
\textbf{\scriptsize Musk1} & {\scriptsize 75.25 $\pm$ 6.89} & {\scriptsize 69.80 $\pm$ 8.98} & {\scriptsize 56.52 $\pm$ 0.70} & {\scriptsize 75.24 $\pm$ 7.11} & {\scriptsize 55.90 $\pm$ 5.09} & {\scriptsize 74.80 $\pm$ 2.88} & {\scriptsize 77.10 $\pm$ 4.50} & {\scriptsize 76.29 $\pm$ 6.76}\tabularnewline
\textbf{\scriptsize Musk2 } & {\scriptsize 83.86 $\pm$ 2.03} & {\scriptsize 77.36 $\pm$ 2.21} & {\scriptsize 84.59 $\pm$ 0.07} & {\scriptsize 83.71 $\pm$ 1.68} & {\scriptsize 94.13 $\pm$ 0.50} & {\scriptsize 95.74 $\pm$ 0.56} & {\scriptsize 89.51 $\pm$ 1.98} & {\scriptsize 83.98 $\pm$ 1.83}\tabularnewline
\textbf{\scriptsize pima-diabetes } & {\scriptsize 76.31 $\pm$ 5.52} & {\scriptsize 70.18 $\pm$ 3.69} & {\scriptsize 71.74 $\pm$ 5.37} & {\scriptsize 76.83 $\pm$ 5.66} & {\scriptsize 72.13 $\pm$ 4.50} & {\scriptsize 71.88 $\pm$ 4.37} & {\scriptsize 76.18 $\pm$ 4.69} & {\scriptsize 74.09 $\pm$ 4.80}\tabularnewline
\textbf{\scriptsize Ecoli } & {\scriptsize 85.40 $\pm$ 5.39} & {\scriptsize 86.92 $\pm$ 3.16} & {\scriptsize 80.37 $\pm$ 5.91} & {\scriptsize 87.18 $\pm$ 4.49} & {\scriptsize 84.52 $\pm$ 5.43} & {\scriptsize 84.52 $\pm$ 5.43} & {\scriptsize 85.40 $\pm$ 5.39} & {\scriptsize 86.31 $\pm$ 6.17}\tabularnewline
\textbf{\scriptsize Glass} & {\scriptsize 49.48 $\pm$ 9.02} & {\scriptsize 48.07 $\pm$ 11.39} & {\scriptsize 15.61 $\pm$ 10.16} & {\scriptsize 50.82 $\pm$ 10.46} & {\scriptsize 59.29 $\pm$ 11.09} & {\scriptsize 60.24 $\pm$ 10.36} & {\scriptsize 49.48 $\pm$ 9.02} & {\scriptsize 54.16 $\pm$ 8.92}\tabularnewline
\textbf{\scriptsize Hill Valley with noise} & {\scriptsize 49.50 $\pm$ 2.94} & {\scriptsize 49.75 $\pm$ 3.40} & {\scriptsize 49.50 $\pm$ 2.94} & {\scriptsize 50.74 $\pm$ 2.88} & {\scriptsize 50.82 $\pm$ 2.93} & {\scriptsize 53.63 $\pm$ 3.77} & {\scriptsize 49.25 $\pm$ 3.39} & {\scriptsize 52.14 $\pm$ 4.21}\tabularnewline
\textbf{\scriptsize Hill Valley w/o noise} & {\scriptsize 51.57 $\pm$ 2.64} & {\scriptsize 50.82 $\pm$ 3.00} & {\scriptsize 51.40 $\pm$ 2.52} & {\scriptsize 51.90 $\pm$ 3.16} & {\scriptsize 51.74 $\pm$ 3.25} & {\scriptsize 52.06 $\pm$ 2.53} & {\scriptsize 51.57 $\pm$ 2.61} & {\scriptsize 52.56 $\pm$ 3.51}\tabularnewline
\textbf{\scriptsize Ionosphere} & {\scriptsize 82.62 $\pm$ 5.47} & {\scriptsize 83.21 $\pm$ 6.42} & {\scriptsize 67.80 $\pm$ 12.65} & {\scriptsize 81.48 $\pm$ 5.42} & {\scriptsize 92.59 $\pm$ 4.71} & {\scriptsize 93.17 $\pm$ 3.06} & {\scriptsize 92.04 $\pm$ 4.37} & {\scriptsize 84.63 $\pm$ 5.02}\tabularnewline
\textbf{\scriptsize Iris } & {\scriptsize 96.00 $\pm$ 4.66} & {\scriptsize 94.67 $\pm$ 4.22} & {\scriptsize 96.67 $\pm$ 3.51} & {\scriptsize 95.33 $\pm$ 5.49} & {\scriptsize 91.33 $\pm$ 6.32} & {\scriptsize 91.33 $\pm$ 6.32} & {\scriptsize 93.33 $\pm$ 7.03} & {\scriptsize 98.00 $\pm$ 3.22}\tabularnewline
\textbf{\scriptsize Isolet } & {\scriptsize 85.15 $\pm$ 0.96} & {\scriptsize 89.06 $\pm$ 0.83} & {\scriptsize 3.85 $\pm$ 0.00} & {\scriptsize 85.58 $\pm$ 0.95} & {\scriptsize 91.83 $\pm$ 0.96} & {\scriptsize 92.97 $\pm$ 0.87} & {\scriptsize 85.15 $\pm$ 0.96} & {\scriptsize 90.68 $\pm$ 0.62}\tabularnewline
\textbf{\scriptsize Letter } & {\scriptsize 64.11 $\pm$ 0.76} & {\scriptsize 59.27 $\pm$ 2.52} & {\scriptsize 24.82 $\pm$ 8.81} & {\scriptsize 64.18 $\pm$ 0.81} & {\scriptsize 58.31 $\pm$ 0.70} & {\scriptsize 56.90 $\pm$ 1.52} & {\scriptsize 64.11 $\pm$ 0.76} & {\scriptsize 67.51 $\pm$ 0.96}\tabularnewline
\textbf{\scriptsize Madelon} & {\scriptsize 58.40 $\pm$ 0.77} & {\scriptsize 59.80 $\pm$ 2.06} & {\scriptsize 50.85 $\pm$ 2.35} & {\scriptsize 58.40 $\pm$ 0.84} & {\scriptsize 55.10 $\pm$ 4.40} & {\scriptsize 60.55 $\pm$ 4.01} & {\scriptsize 53.65 $\pm$ 3.59} & {\scriptsize 58.80 $\pm$ 1.51}\tabularnewline
\textbf{\scriptsize Multiple features} & {\scriptsize 95.35 $\pm$ 1.40} & {\scriptsize 83.40 $\pm$ 2.22} & {\scriptsize 10.90 $\pm$ 1.47} & {\scriptsize 95.15 $\pm$ 1.25} & {\scriptsize 89.05 $\pm$ 2.09} & {\scriptsize 96.05 $\pm$ 1.28} & {\scriptsize 96.40 $\pm$ 0.91} & {\scriptsize 95.25 $\pm$ 1.57}\tabularnewline
\textbf{\scriptsize Sat} & {\scriptsize 79.58 $\pm$ 1.46} & {\scriptsize 81.90 $\pm$ 1.13} & {\scriptsize 69.82 $\pm$ 4.57} & {\scriptsize 79.61 $\pm$ 1.50} & {\scriptsize 85.63 $\pm$ 1.25} & {\scriptsize 86.23 $\pm$ 1.16} & {\scriptsize 79.58 $\pm$ 1.46} & {\scriptsize 83.36 $\pm$ 1.52}\tabularnewline
\textbf{\scriptsize Waveform with noise} & {\scriptsize 80.00 $\pm$ 1.96} & {\scriptsize 80.46 $\pm$ 1.76} & {\scriptsize 72.04 $\pm$ 7.11} & {\scriptsize 80.00 $\pm$ 2.01} & {\scriptsize 84.36 $\pm$ 1.81} & {\scriptsize 84.48 $\pm$ 1.46} & {\scriptsize 80.00 $\pm$ 1.96} & {\scriptsize 81.80 $\pm$ 1.81}\tabularnewline
\textbf{\scriptsize Waveform w/o noise} & {\scriptsize 81.02 $\pm$ 1.33} & {\scriptsize 80.48 $\pm$ 2.03} & {\scriptsize 74.48 $\pm$ 5.23} & {\scriptsize 81.06 $\pm$ 1.35} & {\scriptsize 82.94 $\pm$ 1.62} & {\scriptsize 83.44 $\pm$ 1.72} & {\scriptsize 81.02 $\pm$ 1.33} & {\scriptsize 83.26 $\pm$ 1.59}\tabularnewline
\textbf{\scriptsize Yeast} & {\scriptsize 57.61  $\pm$ 3.01} & {\scriptsize 55.39 $\pm$ 2.33} & {\scriptsize 38.27 $\pm$ 7.89} & {\scriptsize 57.82 $\pm$ 2.69} & {\scriptsize 53.44 $\pm$ 3.96} & {\scriptsize 53.37 $\pm$ 3.94} & {\scriptsize 57.61 $\pm$ 3.01} & {\scriptsize 55.46 $\pm$ 3.23}\tabularnewline
\hline 
\textbf{\footnotesize Average improvement} & {\footnotesize - } & \textbf{\footnotesize -2.3\%} & \textbf{\footnotesize -26.1\% } & \textbf{\footnotesize 0.4\%} & \textbf{\footnotesize 0.2\%} & \textbf{\footnotesize 3.4\%} & \textbf{\footnotesize 0.6\%} & \textbf{\footnotesize 2.3\%}\tabularnewline
\hline 
\textbf{\footnotesize Average rank } & \textbf{\footnotesize 4.71} & \textbf{\footnotesize 5.41} & \textbf{\footnotesize 7.15} & \textbf{\footnotesize 3.97} & \textbf{\footnotesize 4.29} & \textbf{\footnotesize 3.06} & \textbf{\footnotesize 4.59} & \textbf{\footnotesize 2.82}\tabularnewline
\hline 
\multicolumn{7}{c}{} &  & \tabularnewline
\end{tabular}
\end{sidewaystable}

When we compare the average accuracy improvement across all the inducers,
the RPE and DM+AdaBoost were ranked second and third - improving the
plain algorithm by 4\% and 2.1\%, respectively. The Rotation Forest
algorithm is ranked first with 6.4\% improvement. Comparing only the
proposed ensembles according to their average rank as described in
\citep{demsar2006} yielded the following ranking: DM+AdaBoost, RPE,
RSE, DM. The null hypothesis that the RPE, RSE, DM and DM+AdaBoost
algorithms have the same accuracy as the plain algorithm could not
be rejected at confidence level 90\%. Thus, according to the average
accuracy improvement across all the inducers, RPE performs best. However,
according to the average rank, DM+AdaBoost performs best.

\subsection{Discussion }

The results indicate that when a dimensionality reduction algorithm
is coupled with an appropriate inducer, an effective ensemble can
be constructed. For example, the RPE algorithm achieves the best average
improvements when it is paired with the nearest-neighbor and the decision
tree inducers. However, when it is used with the Naïve Bayes inducer,
it fails to improve the plain algorithm. On the other hand, the DM+AdaBoost
ensemble obtains the best average improvement when it is used with
the Naïve Bayes inducer (better than the current state-of-the-art
Rotation Forest ensemble algorithm) and it is less accurate when coupled
with the decision tree and nearest-neighbor inducers. 

Furthermore, using dimensionality reduction as part of a multi-strategy
ensemble classifier improved in most cases the results of the ensemble
classifiers which employed only one of the strategies. Specifically,
the DM+AdaBoost algorithm achieved higher average ranks compared to
the DM and AdaBoost algorithms when the J48 and Naïve Bayes inducers
were used. When the nearest-neighbor inducer was used, the DM+AdaBoost
algorithm was ranked after the DM algorithm and before the AdaBoost
ensemble which was last.

\section{Conclusion and future work \label{sec:Conclusion-and-future}}

In this paper we presented dimensionality reduction as a general framework
for the construction of ensemble classifiers which use a single induction
algorithm. The dimensionality reduction algorithm was applied to the
training set where each combination of parameter values produced a
different version of the training set. The ensemble members were constructed
based on the produced training sets. In order to classify a new sample,
it was first embedded into the dimension-reduced space of each training
set using out-of-sample extension such as the Nyström extension. Then,
each classifier was applied to the embedded sample and a voting scheme
was used to derive the classification of the ensemble. This approach
was demonstrated using three dimensionality reduction algorithms -
Random Projections, Diffusion Maps and Random subspaces. A fourth
ensemble algorithm employed a multi-strategy approach combining the
Diffusion Maps dimensionality reduction algorithm with the AdaBoost
ensemble algorithm. The performance of the obtained ensembles was
compared with the Bagging, AdaBoost and Rotation Forest ensemble algorithms.

The results in this paper show that the proposed approach is effective
in many cases. Each dimensionality reduction algorithm achieved results
that were superior in many of the datasets compared to the plain algorithm
and in many cases outperformed the reference algorithms. However,
when the Naïve Bayes inducer was combined with the Random Subspaces
dimensionality reduction algorithm, the obtained ensemble did not
perform well in some of the datasets. Consequently, a question that
needs further investigation is how to couple a given dimensionality
reduction algorithm with an appropriate inducer to obtain the best
performance. Ideally, rigorous criteria should be formulated. However,
until such criteria are found, pairing dimensionality reduction algorithms
with inducers in order to find the best performing pair can be done
empirically using benchmark datasets. Furthermore, other dimensionality
reduction techniques should be explored. For this purpose, the Nyström
out-of-sample extension may be used with any dimensionality reduction
method that can be formulated as a kernel method \citep{kernel_view_of_DR}.
Additionally, other out-of-sample extension schemes should also be
explored e.g. the Geometric Harmonics \citep{CL_GH06}. Lastly, a
heterogeneous model which combines several dimensionality reduction
techniques is currently being investigated by the authors.

\section*{Acknowledgments}

The authors would like to thank Myron Warach for his insightful remarks.

\bibliographystyle{plain}
\bibliography{refs}

\begin{thebibliography}{10}

\bibitem{UCI}
A.~Asuncion and D.~J. Newman.
\newblock {UCI} machine learning repository, 2007.

\bibitem{Schclar-Neta-detection12}
A.~Averbuch, N.~Rabin, A.~Schclar, and V.~A. Zheludev.
\newblock Dimensionality reduction for detection of moving vehicles.
\newblock {\em Pattern Analysis and Applications}, 15(1):19--27, 2012.

\bibitem{mine-vehicle-wave07}
A.~Averbuch, V.~Zheludev, N.~Rabin, and A.~Schclar.
\newblock Wavelet based detection of moving vehicles.
\newblock {\em International Journal of Wavelets, Multiresolution and
  Information Processing}, http://dx.doi.org/10.1007/s11045-008-0058-z, 2008.

\bibitem{Laplacian03}
M.~Belkin and P.~Niyogi.
\newblock Laplacian eigenmaps for dimensionality reduction and data
  representation.
\newblock {\em Neural Computation}, 15(6):1373--1396, 2003.

\bibitem{KPCA_view04_1}
Y.~Bengio, O.~Delalleau, N.~Le Roux, J.~F. Paiement, P.~Vincent, and M.~Ouimet.
\newblock Learning eigenfunctions links spectral embedding and kernel pca.
\newblock {\em Neural Computation}, 16(10):2197--2219, 2004.

\bibitem{BM2001}
E.~Bingham and H.~Mannila.
\newblock Random projection in dimensionality reduction: applications to image
  and text data.
\newblock In {\em Proceedings of the 7th ACM SIGKDD International Conference on
  Knowledge Discovery and Data Mining (KDD-2001)}, pages 245--250, San
  Francisco, CA, USA, August 26-29 2001.

\bibitem{B85}
J.~Bourgain.
\newblock On lipschitz embedding of finite metric spaces in {Hilbert} space.
\newblock {\em Israel Journal of Mathematics}, 52:46--52, 1985.

\bibitem{Bagging96}
L.~Breiman.
\newblock Bagging predictors.
\newblock {\em Machine Learning}, 24(2):123--140, 1996.

\bibitem{CRT93}
L.~Breiman, J.~H. Friedman, R.~A. Olshen, and C.~J. Stone.
\newblock {\em Classification and Regression Trees}.
\newblock Chapman \& Hall, Inc., New York, 1993.

\bibitem{CRT06}
E.~Cand{\`e}s, J.~Romberg, and T.~Tao.
\newblock Robust uncertainty principles: Exact signal reconstruction from
  highly incomplete frequency information.
\newblock {\em IEEE Transactions on Information Theory}, 52(2):489--509,
  February 2006.

\bibitem{C97}
F.~R.~K. Chung.
\newblock {\em Spectral Graph Theory}.
\newblock {AMS} Regional Conference Series in Mathematics, 92, 1997.

\bibitem{CL_DM06}
R.~R. Coifman and S.~Lafon.
\newblock Diffusion maps.
\newblock {\em Applied and Computational Harmonic Analysis: special issue on
  Diffusion Maps and Wavelets}, 21:5--30, July 2006.

\bibitem{CL_GH06}
R.~R. Coifman and S.~Lafon.
\newblock Geometric harmonics: a novel tool for multiscale out-of-sample
  extension of empirical functions.
\newblock {\em Applied and Computational Harmonic Analysis: special issue on
  Diffusion Maps and Wavelets}, 21:31--52, July 2006.

\bibitem{MDS_94}
T.~Cox and M.~Cox.
\newblock {\em Multidimensional scaling}.
\newblock Chapman \& Hall, London, UK, 1994.

\bibitem{demsar2006}
J.~Demsar.
\newblock Statistical comparisons of classifiers over multiple data sets.
\newblock {\em Journal of Machine Learning Research}, 7:1--30, 2006.

\bibitem{D06}
D.~L. Donoho.
\newblock Compressed sensing.
\newblock {\em IEEE Transactions on Information Theory}, 52(4):1289--1306,
  April 2006.

\bibitem{Hessian02}
D.~L. Donoho and C.~Grimes.
\newblock Hessian eigenmaps: new locally linear embedding techniques for
  high-dimensional data.
\newblock In {\em Proceedings of the National Academy of Sciences}, volume
  100(10), pages 5591--5596, May 2003.

\bibitem{AdaBoost_R2}
H.~Drucker.
\newblock Improving regressor using boosting.
\newblock In D.~H.~Fisher Jr., editor, {\em Proceedings of the 14th
  International Conference on Machine Learning}, pages 107--115. Morgan
  Kaufmann, 1997.

\bibitem{fern2003random}
X.~Z. Fern and C.~E. Brodley.
\newblock Random projection for high dimensional data clustering: A cluster
  ensemble approach.
\newblock In {\em International Conference on Machine Learning (ICML'03)},
  pages 186--193, 2003.

\bibitem{AdaBoost96}
Y.~Freund and R.~Schapire.
\newblock Experiments with a new boosting algorithm. machine learning.
\newblock In {\em Proceedings for the Thirteenth International Conference},
  pages 148--156, San Francisco, 1996. Morgan Kaufmann.

\bibitem{ManhattanNonnegativeMatrixFactorization}
N.~Guan, D.~Tao, Z.~Luo, and J.~Shawe-Taylor.
\newblock Mahnmf: Manhattan non-negative matrix factorization.
\newblock {\em CoRR}, abs/1207.3438, 2012.

\bibitem{WEKA}
M.~Hall, E.~Frank, G.~Holmes, B.~Pfahringer, P.~Reutemann, and I.~H. Witten.
\newblock The weka data mining software: An update.
\newblock {\em SIGKDD Explorations}, 11:1, 2009.

\bibitem{kernel_view_of_DR}
J.~Ham, D.~Lee, S.~Mika, and B.~Scholk{\"o}pf.
\newblock A kernel view of the dimensionality reduction of manifolds.
\newblock In {\em Proceedings of the 21st International Conference on Machine
  Learning (ICML'04)}, pages 369--376, New York, NY, USA, 2004.

\bibitem{Wakin07}
C.~Hegde, M.~Wakin, and R.~G. Baraniuk.
\newblock Random projections for manifold learning.
\newblock In {\em Neural Information Processing Systems (NIPS)}, December 2007.

\bibitem{Hein05}
M.~Hein and Y.~Audibert.
\newblock Intrinsic dimensionality estimation of submanifolds in {Euclidean}
  space.
\newblock In {\em Proceedings of the 22nd International Conference on Machine
  Learning}, pages 289--296, 2005.

\bibitem{RandomSubspaceDecisionForest}
T.~K. Ho.
\newblock The random subspace method for constructing decision forests.
\newblock {\em IEEE Transaction on Pattern Analysis and Machine Intelligence},
  20(8):832--844, 1998.

\bibitem{PCA}
H.~Hotelling.
\newblock Analysis of a complex of statistical variables into principal
  components.
\newblock {\em Journal of Educational Psychology}, 24:417--441, 1933.

\bibitem{JL98}
L.~O. Jimenez and D.~A. Landgrebe.
\newblock Supervised classification in high-dimensional space: geometrical,
  statistical and asymptotical properties of multivariate data.
\newblock {\em IEEE Transactions on Systems, Man and Cybernetics, Part C:
  Applications and Reviews,}, 28(1):39--54, February 1998.

\bibitem{JL84}
W.~B. Johnson and J.~Lindenstrauss.
\newblock Extensions of {Lipshitz} mapping into {Hilbert} space.
\newblock {\em Contemporary Mathematics}, 26:189--206, 1984.

\bibitem{MDS64}
J.~B. Kruskal.
\newblock Multidimensional scaling by optimizing goodness of fit to a nonmetric
  hypothesis.
\newblock {\em Psychometrika}, 29:1--27, 1964.

\bibitem{Kuncheva:2004}
L.~I. Kuncheva.
\newblock {\em Combining Pattern Classifiers: Methods and Algorithms}.
\newblock Wiley-Interscience, 2004.

\bibitem{EnsembleDiversity04}
L.~I. Kuncheva.
\newblock Diversity in multiple classifier systems (editorial).
\newblock {\em Information Fusion}, 6(1):3--4, 2004.

\bibitem{Lafon06datafusion}
S.~Lafon, Y.~Keller, and R.~R. Coifman.
\newblock Data fusion and multicue data matching by diffusion maps.
\newblock {\em IEEE Transactions on Pattern Analysis and Machine Intelligence},
  28:1784--1797, 2006.

\bibitem{Leigh02-1}
W.~Leigh, R.~Purvis, and J.~M. Ragusa.
\newblock Forecasting the nyse composite index with technical analysis, pattern
  recognizer, neural networks, and genetic algorithm: a case study in romantic
  decision support.
\newblock {\em Decision Support Systems}, 32(4):361--377, 2002.

\bibitem{MultitrainingLiATL06}
J.~Li, N.~M. Allinson, D.~Tao, and X.~Li.
\newblock Multitraining support vector machine for image retrieval.
\newblock {\em IEEE Transactions on Image Processing}, 15(11):3597--3601, 2006.

\bibitem{LLTY00}
M.~Linial, N.~Linial, N.~Tishby, and G.~Yona.
\newblock Global self-organization of all known protein sequences reveals
  inherent biological signatures.
\newblock {\em Journal of Molecular Biology}, 268(2):539--556, May 1997.

\bibitem{Luo2012}
Y.~Luo, D.~Tao, B.~Geng, C.~Xu, and S.~Maybank.
\newblock Manifold regularized multi-task learning for semi-supervised
  multi-label image classification.
\newblock {\em IEEE Transaction on Image Processing}, 22(2):523--536, 2013.

\bibitem{Mangiameli04}
P.~Mangiameli, D.~West, and R.~Rampal.
\newblock Model selection for medical diagnosis decision support systems.
\newblock {\em Decision Support Systems}, 36(3):247--259, 2004.

\bibitem{KappaError97}
D.~D. Margineantu and T.~G. Dietterich.
\newblock Pruning adaptive boosting.
\newblock In {\em Proceedings of the 14th International Conference on Machine
  Learning}, pages 211--218, 1997.

\bibitem{N28}
E.~J. {Nystr{\"o}m}.
\newblock {\"U}ber die praktische aufl{\"o}sung von linearen
  integralgleichungen mit anwendungen auf randwertaufgaben der
  potentialtheorie.
\newblock {\em Commentationes Physico-Mathematicae}, 4(15):1--52, 1928.

\bibitem{Opitz}
D.~Opitz and R.~Maclin.
\newblock Popular ensemble methods: An empirical study.
\newblock {\em Journal of Artificial Intelligence Research}, 11:169 to 198,
  1999.

\bibitem{Plastria}
F.~Plastria, S.~Bruyne, and E.~Carrizosa.
\newblock Dimensionality reduction for classification.
\newblock {\em Advanced Data Mining and Applications}, 1:411--418, 2008.

\bibitem{Polikar}
R.~Polikar.
\newblock "ensemble based systems in decision making.
\newblock {\em IEEE Circuits and Systems Magazine}, 6:21 t o 45, 2006.

\bibitem{C45Quinlan}
R.~R. Quinlan.
\newblock {\em C4.5: programs for machine learning}.
\newblock Morgan Kaufmann Publishers Inc., 1993.

\bibitem{RotationForest2006}
J.~J. Rodriguez, L.~I. Kuncheva, and C.~J. Alonso.
\newblock Rotation forest: A new classifier ensemble method.
\newblock {\em IEEE Transactions on Pattern Analysis and Machine Intelligence},
  28(10):1619--1630, 2006.

\bibitem{Rokach08}
L.~Rokach.
\newblock Mining manufacturing data using genetic algorithm-based feature set
  decomposition.
\newblock {\em International Journal of Intelligent Systems Technologies and
  Applications}, 4(1/2):57--78, 2008.

\bibitem{Rokach2009}
L.~Rokach.
\newblock Taxonomy for characterizing ensemble methods in classification tasks:
  A review and annotated bibliography.
\newblock {\em Computational Statistics \& Data Analysis}, In Press, Corrected
  Proof:--, 2009.

\bibitem{RandSubSpaceEnsemble04}
N.~Rooney, D.~Patterson, A.~Tsymbal, and 10~February~2004 S.~Anand.
\newblock Random subspacing for regression ensembles.
\newblock Technical report, Department of Computer Science, Trinity College
  Dublin, Ireland, 2004.

\bibitem{LLE00}
S.~T. Roweis and L.~K. Saul.
\newblock Nonlinear dimensionality reduction by locally linear embedding.
\newblock {\em Science}, 290:2323--2326, December 2000.

\bibitem{SchclarDiffusionFrame}
A.~Schclar.
\newblock A diffusion framework for dimensionality reduction.
\newblock In {\em Soft Computing for Knowledge Discovery and Data Mining
  (Editors: O. Maimon and L. Rokach)}, pages 315--325. Springer, 2008.

\bibitem{SchclarDetection2010}
A.~Schclar, A.~Averbuch, N.~Rabin, V.~Zheludev, and K.~Hochman.
\newblock A diffusion framework for detection of moving vehicles.
\newblock {\em Digital Signal Processing}, 20:111--122, January 2010.

\bibitem{SchclarICEIS09}
A.~Schclar and L.~Rokach.
\newblock Random projection ensemble classifiers.
\newblock In {\em Lecture Notes in Business Information Processing, Enterprise
  Information Systems 11th International Conference Proceedings (ICEIS'09)},
  pages 309--316, Milan, Italy, May 2009.

\bibitem{SchclarRecSys09}
A.~Schclar, A.~Tsikinovsky, L.~Rokach, A.~Meisels, and L.~Antwarg.
\newblock Ensemble methods for improving the performance of neighborhood-based
  collaborative filtering.
\newblock In {\em RecSys}, pages 261--264, 2009.

\bibitem{KPCA98}
B.~Sch\"{o}lkopf, A.~Smola, and K.~R. Muller.
\newblock Nonlinear component analysis as a kernel eigenvalue problem.
\newblock {\em Neural Computation}, 10(5):1299--1319, 1998.

\bibitem{KPCA_book02}
B.~Sch\"{o}lkopf and A.~J. Smola.
\newblock {\em Learning with Kernels}.
\newblock MIT Press, Cambridge, MA, 2002.

\bibitem{AdaBoost_RT}
D.~P. Solomatine and D.~L. Shrestha.
\newblock Adaboost.rt: A boosting algorithm for regression problems.
\newblock In {\em Proceedings of the IEEE International Joint Conference on
  Neural Networks}, pages 1163--1168, 2004.

\bibitem{GeometricMean_for_SubspaceSelection}
D.~Tao, X.~Li, X.~Wu, and J.~S. Maybank.
\newblock Geometric mean for subspace selection.
\newblock {\em IEEE Transactions on Pattern Analysis and Machine Intelligence},
  31(2):260--274, February 2009.

\bibitem{Asymmetric_Bagging_Image_Retrieval}
D.~Tao, X.~Tang, X.~Li, and X.~Wu.
\newblock Asymmetric bagging and random subspace for support vector
  machines-based relevance feedback in image retrieval.
\newblock {\em IEEE Transactions on Pattern Analysis and Machine Intelligence},
  28(7):1088--1099, July 2006.

\bibitem{ISO00}
J.~B. Tenenbaum, V.~de~Silva, and J.~C. Langford.
\newblock A global geometric framework for nonlinear dimensionality reduction.
\newblock {\em Science}, 290:2319--2323, December 2000.

\bibitem{BaggEnsembleApp03}
G.~Valentini, M.~Muselli, and F.~Ruffino.
\newblock Bagged ensembles of svms for gene expression data analysis.
\newblock In {\em Proceeding of the International Joint Conference on Neural
  Networks - IJCNN}, pages 1844--1849, Portland, OR, USA, July 2003. Los
  Alamitos, CA: IEEE Computer Society.

\bibitem{SVM99}
V.~N. Vapnik.
\newblock {\em The Nature of Statistical Learning Theory (Information Science
  and Statistics)}.
\newblock Springer, November 1999.

\bibitem{multiboosting:a:Webb00}
G.~I. Webb.
\newblock Multiboosting: A technique for combining boosting and wagging.
\newblock In {\em Machine Learning}, pages 159--196, 2000.

\bibitem{multi-strategyensemble:Webb04}
G.~I. Webb and Z.~Zheng.
\newblock Multi-strategy ensemble learning: Reducing error by combining
  ensemble learning techniques.
\newblock {\em IEEE Transactions on Knowledge and Data Engineering}, 16:2004,
  2004.

\bibitem{BaggEnsembleApp06}
Z.~Yang, X.~Nie, W.~Xu, and J.~Guo.
\newblock An approach to spam detection by naive bayes ensemble based on
  decision induction.
\newblock In {\em Proceedings of the Sixth International Conference on
  Intelligent Systems Design and Applications (ISDA'06)}, 2006.

\bibitem{PatchAlignment_forDimensionalityReduction}
T.~Zhang, D.~Tao, X.~Li, and J.~Yang.
\newblock Patch alignment for dimensionality reduction.
\newblock {\em IEEE Transactions on Knowledge and Data Engineering},
  21(9):1299--1313, September 2009.

\bibitem{LocalTanSpAnal02}
Z.~Zhang and H.~Zha.
\newblock Principal manifolds and nonlinear dimension reduction via local
  tangent space alignment, 2002.

\bibitem{Manifold_elastic_net}
T.~Zhou, D.~Tao, and X.~Wu.
\newblock Manifold elastic net: a unified framework for sparse dimension
  reduction.
\newblock {\em Data Mining and Knowledge Discovery}, 22(3):340--371, May 2011.

\end{thebibliography}

\end{document}